\documentclass{article}

    \PassOptionsToPackage{numbers, compress}{natbib}

     \usepackage[preprint]{neurips_2020}

\usepackage[preprint]{neurips_2020}

\usepackage[utf8]{inputenc} 
\usepackage[T1]{fontenc}    
\usepackage{hyperref}       
\usepackage{url}            
\usepackage{booktabs}       
\usepackage{amsfonts}       
\usepackage{nicefrac}       
\usepackage{microtype}      
\usepackage{amsmath,graphicx}

\usepackage{mathtools}

\usepackage{enumitem}
\newtheorem{theorem}{Theorem}
\newtheorem{defn}{Definition}
\usepackage[ruled,vlined]{algorithm2e}

\newcommand\myeq{\stackrel{\mathclap{\normalfont\mbox{{\tiny i.i.d}}}}{\sim}}
\usepackage{caption}
\usepackage{subcaption}

\title{Bayesian Nonparametric Modelling for Model-Free Reinforcement Learning in LTE-LAA and Wi-Fi Coexistence}

\author{
  Po-Kan Shih \\
  School of Electrical, Computer and Energy Engineering\\
  Arizona State University\\
  Tempe, AZ 85281 \\
  \texttt{pshih3@asu.edu} \\
  \AND
  Bahman Moraffah \\
  School of Electrical, Computer and Energy Engineering\\
  Arizona State University\\
  Tempe, AZ 85281 \\
  \texttt{bahman.moraffah@asu.edu} \\
}

\begin{document}

\maketitle

\begin{abstract}
With the arrival of next generation wireless communication, a growing number of new applications like internet of things, autonomous driving systems, and drone are crowding the unlicensed spectrum. Licensed network such as the long-term evolution (LTE) also comes to the unlicensed spectrum for better providing high-capacity contents with low cost. However, LTE was not designed to share resources with others. Previous solutions usually work on fixed scenarios. This work features a Nonparametric Bayesian reinforcement learning algorithm to cope with the coexistence between Wi-Fi and LTE licensed assisted access (LTE-LAA) agents in 5 GHz unlicensed spectrum. The coexistence problem is modeled as a decentralized partially-observable Markov decision process (Dec-POMDP) and Bayesian inference is adopted for policy learning with nonparametric prior to accommodate the uncertainty of policy for different agents. A fairness measure is introduced in the reward function to encourage fair sharing between agents. Variational inference for posterior model approximation is considered to make the algorithm computationally efficient. Simulation results demonstrate that this algorithm can reach high value with compact policy representations in few learning iterations.
\end{abstract}

\section{Introduction}
\label{sec: intro}

Spectrum coexistence has been a popular problem to be studied as the wireless networks are proceeding to the fifth generation (5G). 5G standard aims to augment the transmission throughput for different classes of usage such as ultra-dense, scalable, or highly-customizable connections. In order to achieve this while not raising the price significantly for terminal customers, it is essential for the cellular network operators to seek more spectrum resources since current licensed spectrum is limited and costly. Offloading to the unlicensed spectrum generates two explicit advantages: access flexibility for abrupt transmission requests and cost efficiency. There are two unlicensed spectrum centered at 2.4GHz and 5GHz to provide free access without need of permission. However, different types of wireless network have crowded the unlicensed spectrum, including Wi-Fi, bluetooth, various internet of things (IoTs); other novel applications, such as autonomous driving systems, radar, drone, are still arriving. According to the Cisco annual internet report \cite{cisco}, the amount of Wi-Fi hotspots is expected to be up to 628 million, and the population of cellular network subscribers will grow to 5.7 billion by year 2023. Such numbers imply that there is potentially infinite amount of wireless devices continuously coming and leaving the spectrum. These networks possess heterogeneous properties such as quality of services (QoS), protocols, bandwidth requirements, and access timing. The nature that their spectrum access requirements are not even similar toughens the coordination among networks. For instance, cellular networks such as the Long-Term Evolution (LTE) demand stable and exclusive channel resource for its high-quality, heavy-load content, whereas countless cloud-integrated IoTs devices usually perform their transmission in a frequent and burst-like way. Such problem has motivated the study for spectrum coexistence techniques \cite{bg_5Gunlincesed,bg_HAN201653,bg_wifiLTEDC}.


In this article, a fair spectrum coexistence scheme between the Wi-Fi and LTE networks in the 5 GHz unlicensed band is considered. The interaction between wireless nodes and the spectrum was framed as an infinite-horizon decentralized partially observable Markov decision process (Dec-POMDP). A cumulative reward function is proposed to reflect the quality of sequential decision on the contention for limited spectrum resource. Our algorithm features model-free, off-policy reinforcement learning to learn policy for each agent from trajectories collected with behavior policy. In addition, Bayesian inference with nonparametric prior for policy learning is adopted to accommodate uncertain policy representations for heterogeneous types of wireless nodes. Variational inference turns the posterior inference into optimization problem in consideration of high-complexity model complexity and scalability. To our best knowledge, this is the first work on spectrum coexistence for LTE and Wi-Fi networks utilizing Bayesian reinforcement learning with variable policy representations.

The rest of this article is organized as follows. Previous studies in spectrum coexistence and current standardization are reviewed in section 2. Section 3 covers the concept of Bayesian inference, the nonparametric model utilized in our algorithm, and its applications in signal processing. In the succeeding section the proposed algorithm is introduced. We first exhibits our approach of modelling a spectrum coexistence scenario as a decentralized partially-observable Markov decision process (Dec-POMDP), then elaborates the iterative algorithms of posterior approximation for policy parameters. Performance evaluation is demonstrated with discussion in the next section. The final Section summarizes our contributions and delivers conclusions as well as some future extensions.

\section{Related Works}
\label{sec: related_work}

Spectrum coexistence is crucial for all wireless devices in unlicensed spectrum. Some industrial solutions have been proposed and employed in current wireless standards \cite{LTEWiFisurvey}. IEEE 802.11 Wi-Fi stansards adopt the carrier sense multiple access/collision avoidance (CSMA/CA) to tackle the coexistence within Wi-Fi networks or with other networks in unlicensed spectrum. CSMA/CA coordinates Wi-Fi nodes in a non-cooperative manner, which involves listen-before-talk (LBT) mechanism to avoid collisions. LBT requests the node (either access point or user equipment) to ensure the channel is clear by performing spectrum sensing (listen) before transmission (talk) commences. By assessing channel before transmission, each node will suspend its transmission until the channel is sensed idle for a duration, which means even if the channel is idle, the node still will not be permitted to occupy the channel immediately. The reason of waiting for an extra idle duration is that there may be multiple nodes waiting for the same channel, and all waiting nodes start their transmission immediately after idle sensing results will cause severe collision. To avoid coincident transmission, CSMA/CA mandates node to execute back-off sensing for extra time slots, and transmission is allowed to start only when the back-off sensing result is idle for the extra time slots. With stochastic numbers of time slots for different nodes, probability of coincident transmission is reduced. On the other hand, LTE operators exhibit their interest in unlicensed spectrum with the growing demand for high-quality, low-latency services. Some methods have been proposed for LTE networks to operate in unlicensed spectrum. Based on the coordination mechanism, LBT and almost blank sub-frame (ABS) are two main categories. Unlike LBT mechanism, ABS employs duty-cycle for LTE nodes to share spectrum with other networks. LTE node actively interrupts its transmission for other networks to compete for the channel periodically, and the interrupt period depends on the measure of channel activity. Compared to LBT methods, ABS-based methods sense channel less frequently and need less modification in current LTE framework in exchange of less efficiency. LBT-based methods can be deployed in areas like Europe and Japan where channel assessment before transmission is compulsory, although it requires some modification in current LTE framework. ABS-based methods can be implemented easily in areas without requirement of assessment before transmission, such as United States, China, Korea, and India. In standardization effort, the LTE-unlicensed (LTE-U) encodes ABS-based method in 3GPP LTE release 10. Among all methods, the LTE-licensed assisted access (LTE-LAA) is one of the most competitor because its framework is similar to Wi-Fi and thus can be adopted in all regions in the world. The LTE-LAA is LBT-based, and has been standardized in 3GPP LTE release 13 to offload downlink traffic to the unlicensed spectrum in 5 GHz \cite{LTEWiFisurvey}.

Academic has proposed some solutions as well. Kota in \cite{kota} and Sodagari in \cite{exmethod_6503914} proposed a joint waveform design to suppress the co-channel interference between radar and multi-input, multi-output communication systems. Their multi-objective cost functions involve all channel users to search for the jointly optimal waveform. Nevertheless, such approaches lack of capability of generalization, and need to start over for any condition change or optimize every possible configuration in advance and store them in memory. Furthermore, they are sensitive to noisy information; information such as number of spectrum users, user types, or bandwidth allocation, must be completely correct or the optimization result will be useless, which is infeasible in real-world applications. In the last decade, reinforcement learning (RL) is becoming a popular approach to the spectrum coexistence problem. Conventional Q-learning was adopted in \cite{Q_8645080} to dynamically manage transmission power of radar and communication systems for a joint radar-communication receiver. \cite{Q_fairCoex,Q_sensors,8288850} considered the coexistence between LTE-U and Wi-Fi networks, applying Q-learning to learn the optimal active duration in duty-cycle for LTE-U agents for maximizing throughputs conditional on fair spectrum sharing. In \cite{RL_8378616} the author modeled the radar tracking as Markov decision process and searched the optimal linear frequency modulation for different target and interference conditions with policy iterations. In addition to above methods, an analytical model was proposed in \cite{9093212} to measure the throughput of Wi-Fi and LTE-LAA agents and multi-armed bandit algorithm was utilized to adjust the contention window for throughput maximization subject to fairness constraint. These solutions work well with full observability of the spectrum activity and maintain predefined Q-tables which include all possible state-action combinations; the learned policies struggle in stochastic and non-stationary spectrum dynamics, and troubles emerge as the scale of problem enlarging. On the other hand, some solutions are under assumption of partially-observable environments to overcome above issues. \cite{6482133,6417543,6415750,7249182} formulated partially-observable Markov decision processes (POMDPs) for wireless transceivers and optimized the transmitting efficiency in noisy spectrum. Gaussian process was adopted in \cite{9064881} for a time-series POMDP model in consideration of correlation between channels to approximate the Q-functions in Q-table. The author of \cite{8646702} proposed a dynamic Q-dictionary which allows adding non-predetermined state-action combination in the learning process. Secondary users of cognitive radios utilized Q-learning to seek the most efficient strategies for locating the clear channels in different spectrum configurations in \cite{Q_7925694,Q_5986378,Q_dynchanselCR}. Except conventional learning methods, deep learning is also grasping attentions. \cite{9069218} combined a model-free decentralized deep Q-learning with model-based Q-learning to mutually compensate the imperfect results from each other as well as accelerate the learning process. Albeit deep Q-learning can handle partial observability, the Q-table still needs to be defined in advance and each Q-function is approximated by a neural network (NN), which means a significant amount of NNs and each NN requires a great bunch of data to train, let alone additional regularization in avoidance of overfitting.
\section{Nonparametric Bayesian Reinforcement Learning}
\label{sec: PS}

Unlike maximum likelihood estimation, Bayesian inference maintains distributions over latent variables in a model, provides a way to encode prior belief in the distributions, and infers the posteriors using data drawn from the model. Denote $z$ as the desired variable in the model of interest, and $\mathcal{D}=(x_1,x_2,...,x_N)$ as the collected dataset, the Bayesian inference updates the belief about $z$ through Bayes rule
\begin{equation}
    p(z|\mathcal{D})=\frac{p(\mathcal{D}|z)p(z)}{p(\mathcal{D})}
\end{equation}
The main difference that distinguishes Bayesian inference from maximum likelihood estimation lies on the representation of $z$. Bayesian approaches regard unknown value $z$ as random variable; the prior distribution $p(z)$ encrypts belief about how the value of $z$ will distribute before observing the dataset. Since each $z$ value represents an unique estimate of the true model, the distribution over $z$ determines the estimation about the true model. After dataset $\mathcal{D}$ is collected from the true model, we can utilize it to update $p(z)$ to the posterior $p(z|\mathcal{D})$. Likelihood function $p(\mathcal{D}|z)$ represents the distribution (usually a parametric model) over true model and is controlled by $z$, indicating the probability of observing $\mathcal{D}$ conditional on some $z$; the denominator $p(\mathcal{D})$ is the marginal likelihood over $\mathcal{D}$ and is obtained by marginalizing $z$ over its sample space.

The prior distributions in Bayesian inference can be categorized as parametric or nonparametric models. Parametric models have fixed structure and are straightforward. They are often utilized when the distribution family over the true model can be defined clearly. Nonparametric models, on the other hand, generalize the parametric models to infinite dimension to handle a broader range of problems. The objective of adopting nonparametric models is to reserve flexibility of adjusting model structures. In general, parametric models function without problem when the structure of true model is definite and all critical information is available in advance. However, they impose strong prior assumption on the model, which is not the case in most real-world applications. For example, in an inference for a Gaussian mixture model (GMM), if the number of mixture components is known, the prior model can be defined precisely and the inference can be accomplished efficiently and accurately. But if such information is noisy or unavailable, parametric methods may need to perform inference for different mixture configurations, and incorporate extra algorithms like cross validation to extract the optimal result, which causes the methods inefficient and burdensome. Nonparametric methods crack such issue with single algorithm. They cast the structure uncertainty as another latent variable, loosening the limitation of parametric models. By incorporating infinite possibility over model structure, nonparametric methods enable the learning agent to learn model variables and structure simultaneously, thus can well fit inference problems with insufficient prior information.

\subsection{Dirichlet and Stick-Breaking Processes}

Dirichlet process (DP) is a widely-used nonparametric model for discrete distributions. It generalizes the Dirichlet distribution and was first proposed by Ferguson \cite{ferguson1973}. Denote $G_0$ as the base distribution over space $\Theta$ and $\alpha$ as some positive real value, $\theta_1, \theta_2,...$ are drawn i.i.d from $\Theta$ with corresponding probability $p_1,p_2,...$ from $G_0$, a random measure $G$ is represented as discrete distribution with infinitely countable components,
\begin{equation*}
    G=\sum_{i=1}^{\infty}p_i\delta_{\theta_i}, \quad \sum_{i=1}^{\infty}p_i=1,
\end{equation*}
where $\delta$ is the Dirac delta function. By definition, $G$ is distributed according to $\text{DP}(\alpha, G_0)$ if for arbitrary finite measurable partition $(A_1,...,A_n)$ over $\Theta$, the vector of random measure $G(A_1),...,G(A_n)$ follows Dirichlet distribution,
\begin{equation*}
    (G(A_1),...,G(A_n))\sim \text{Diri}(\alpha G_0(A_1),...,\alpha G_0(A_n))
\end{equation*}
The stick-breaking process offers a straightforward approach to construct $G$. The process to generate the probability weights of $G$ is analogous to breaking an unit-length stick into an infinite sequence of pieces. Imagine a stick with length $1$ initially, we break a portion $V_1$ off from the stick at first step. Then a portion $V_2$ is taken off from the remaining part of the stick (with length $1-V_1$) at second step. As the process proceeds, at step $i$ a portion $V_i$ will be broken off from the leftover of the stick. The broken portions are drawn from one or more Beta distributions. Given a random variable $V$ with beta distribution $\text{Beta}(1,\alpha)$, and point mass $\theta_1,\theta_2,...$ drawn from $G_0$, the random probability weights $(p_1,p_2,...)$ in $G$ can be constructed through an unbounded process:
\begin{equation}
    \begin{aligned}
    &\begin{cases}
    \theta_i|G_0\myeq G_0 \\
    V_i|\alpha \myeq \text{Beta}(1,\alpha)
    \end{cases}, i = 1,2,... \\
    & p_1 = V_1, \  p_i= V_i\prod_{j=1}^{i-1}(1-V_j) \text{ for }i>1 \\
    & G=\sum_{i=1}^{\infty}p_i\delta_{\theta_i}
    \end{aligned}
\label{equ: SB}
\end{equation}
The stick-breaking process can construct the random measure $G$ fast and guarantee $p_i$ sum to $1$. The distribution over random $p_i$ is also known as the $\text{GEM}(\alpha)$ distribution \cite{pitman2002GEM}.

Bayesian nonparametric modeling has been exploited in signal processing, especially when the data pattern is uncertain in advance. The author of \cite{DBLPBahman, moraffah2019random, moraffah2018dependent, moraffah2019inference, moraffah2019nonparametric, moraffah2019tracking,moraffah2021clutter} inferred the unbounded number of objects and characteristics of each object together using a dependent Dirichlet process in radar tracking scenarios. Hierarchical Dirichlet process was also adopted in \cite{moraffah2019use, moraffah2019inference} to track multiple time-varying objects. Guo et al. in \cite{guo2018explaining} employed Dirichlet process mixture model to extract insights from deep neural networks, which assists understanding and interpretation of machine learning models, and demonstrated its ability to generalize to different machine learning frameworks. A variance Gamma process was proposed in \cite{pmlr-zhang18j} to encode probabilistic assumptions in the prior model, interpreting sparse and discrete data points in time-series data better than typical machine learning algorithms, which generally yield smooth functions. The author of \cite{DBLPPolatkan} adopted Beta-Bernoulli process for the prior model to learn an unbounded set of visual recurring patterns from data, and augmented image resolution from low-resolution images with the learned set.

\subsection{Decentralized Partially-Observable Markov Decision Process}
\label{sec: decPOMDP}

The decentralized partially-observable Markov decision processes (Dec-POMDPs) generalize the multi-agent Markov decision processes when the states are partially-observable to the agents and each agent performs its own reinforcement learning without cooperation or information exchange. A Dec-POMDP model can be represented by the tuple $\langle\mathcal{N}, \mathcal{A}, \mathcal{S}, \mathcal{O}, T, \Omega, R, \gamma\rangle$ \cite{POMDPbook,RLbook}. $\mathcal{N}=\{1,...,N\}$ is the finite set of agent indices. The state set $\mathcal{S}$ includes all global states $s$ of the model. $\mathcal{A}=\bigotimes_n\mathcal{A}_n$ and $\mathcal{O}=\bigotimes_n\mathcal{O}_n$ correspond to the sets of joint actions and observations, respectively, where $\mathcal{A}_n$ and $\mathcal{O}_n$ are local action and observation sets of agent $n$. In each state, a joint action $\vec{a}=\left\{a_n\right\}_{n=1}^N \in\mathcal{A}$ formed with local actions $a_n\in\mathcal{A}_n$ is performed, and a joint observation $\vec{o}=\left\{o_n\right\}_{n=1}^N \in\mathcal{O}$ is received by the agents, despite agent $n$ only has access to $o_n$. $T(s',\vec{a},s)=\Pr(s'|\vec{a},s)$ $\forall s,s'\in\mathcal{S}$, $\vec{a}$ denotes the state transition probability given state and joint action. $R:\mathcal{S}\times\mathcal{A}\rightarrow\mathbb{Z}$ represents the immediate reward function which yields a global real value $r$ to all agents after $\vec{a}$ is performed in state $s$. The reward function encrypts the core objective of the learning agents, that is, the prior concern for the system; different reward functions lead to different learned results. $\Omega(\vec{o})=\Pr(\vec{o}|s,\vec{a})$ expresses the probability distribution over joint observations when performing action $\vec{a}$ and arriving at state $s$. $\gamma$ is a predefined positive real constant between $[0,1)$ which weights the importance of future rewards against current reward.

\subsection{Bayesian Reinforcement Learning}

Bayesian reinforcement learning applies Bayesian inference to the variable estimation in reinforcement learning, placing prior distributions over the desired variables and calculate the posterior distributions. In this work we adopt policy-based learning, which learns policy directly without knowledge about the underlying environment. The variables to be estimated are policy parameters. Denote $\Theta$ the parameters of policy and $\mathcal{D}$ the dataset from interaction with the environment, the Bayesian policy learning infers the posterior distributions over policy parameters given prior and data:
\begin{equation*}
    p(\Theta|\mathcal{D})=\frac{p(\mathcal{D}|\Theta)p(\Theta)}{p(\mathcal{D})}\propto p(\mathcal{D}|\Theta)p(\Theta).
\label{equ: bayes}
\end{equation*}
$p(\Theta)$ is the prior distribution indicating belief about $\Theta$ before learning process starts. By applying distribution over $\Theta$, it is convenient to impose auxiliary constraints to avoid premature convergence, and quantify our confidence about $\Theta$. By interacting with the environment, data is collected and fed into likelihood $p(\mathcal{D}|\Theta)$ to update the posterior $p(\Theta|\mathcal{D})$. During the iterations of learning process, the posterior distribution obtained at current iteration may serve as the prior distribution for next iteration. As learning process proceeds, the convergence of $p(\Theta)$ is guaranteed. Bayesian learning provides a faster and simpler alternative to deep learning since the presence of prior distribution intrinsically biases the data-driven only likelihood in avoidance of overfitting so that additional regularization algorithm is not necessary. Prior knowledge carried in prior distributions also mitigates the desire for large dataset to match the performance of deep learning.
\section{Proposed Method}
\label{sec: method}

In this section the underlying coexistence scenario between LTE and Wi-Fi networks is first presented, then a Dec-POMDP model for it is formulated, including a cumulative reward function to reflect the continuous spectrum variation. The prior distributions incorporate the nonparametric model introduced in previous section. Variational inference is exploited to cast the posterior calculation to optimization problem; its analytical equations will be articulated.
\subsection{Signal Model}
\begin{figure}
\centerline{\includegraphics[width=1\linewidth]{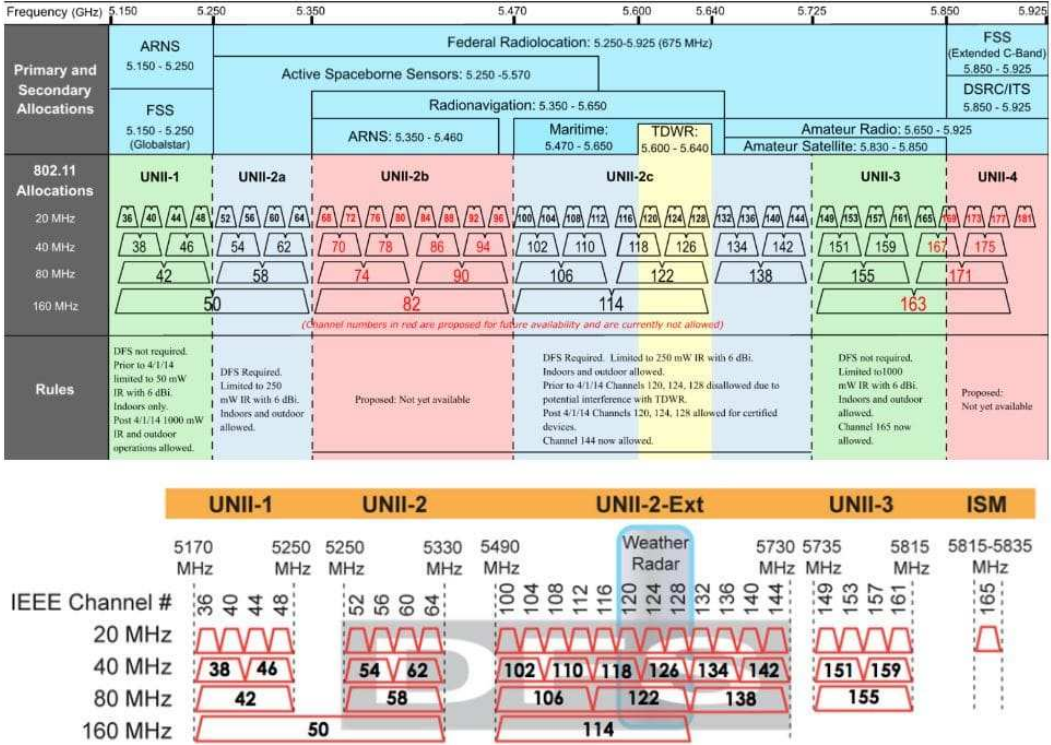}}
	\caption[5G Spectrum Usage]{Illustration for current frequency allocation in 5 GHz spectrum\footnotemark.}
	\label{fig:5GHz}
\end{figure}
The 5 GHz unlicensed spectrum spans approximately 600 MHz and is divided into non-overlapping channels, which current channel allocation is illustrated in \ref{fig:5GHz}. Each unit channel has bandwidth of 20 MHz. If the spectrum is not crowded, wireless nodes are allowed to utilize channels of larger bandwidth (40, 80, or 160 MHz), which consists of multiple consecutive unit channels. In our scenario, each wireless node can be either LTE evolved node B (eNB) or Wi-Fi access point (AP). Due to the property of free access, it is impractical to centralize the management of heterogeneous nodes in the spectrum in single control center; information exchange between heterogeneous networks also suffers trouble due to the divergent protocols and extra costs. Thus a practical spectrum coexistence scheme should consider decentralization, which means communication between nodes is minimized and each node optimizes its own spectrum accessing policy independently. Additionally, nodes like user equipment only possess limited spectrum sensing capability and obtains noisy information about the spectrum. Without access to the global configuration, nodes determine the action to perform according to the sufficient statistics of past observations and actions, which is termed as belief or decision state in some documents. Given aforementioned situations, we select Dec-POMDP model to properly express our spectrum coexistence scenario.

\footnotetext{From \url{https://www.wlanpros.com/5ghz-frequency-allocations-2/}.}

It is worth to note that the unbounded prior for policy learning should be considered. The license-free property allows every node to come and leave the spectrum unrestrictedly, albeit each time there is only a finite set of nodes has the opportunity to occupy the spectrum. Hence we should not expect the amount of potential nodes is bounded and known to learning agents. The policy learning must consider interactions with uncertain number of coexisting nodes, thus nonparametric models which can accommodate infinite policy representations are more appropriate than parametric models. In addition, fair spectrum sharing is another factor crucial to the coexisting nodes and worth more attention. The LTE transmission is frame-based. Each LTE data frame includes multiple sub-frames, where each sub-frame spends 1 ms. The number of sub-frames in one data frame is determined by the access priority of the node \cite{3gpp.36.213}. Wi-Fi, on the other hand, is packet-based. Each Wi-Fi transmission contains only one packet with fixed size. Although the frame aggregation can enhances airtime efficiency by combining multiple packets in single transmission \cite{802.11n_cisco}, we will not consider it here for simplicity. The different frameworks of data frame cause LTE transmission more enduring then Wi-Fi, which makes LTE nodes easier dominate in time domain and thus winner keeps winning, expelling Wi-Fi nodes from competition. If only the performance of the whole spectrum is considered, the learning process may tend to sacrifice weaker nodes for dominant nodes, which is what we aim to avoid. To achieve fairness, we incorporate the most commonly-utilized Jain's fairness indicator \cite{jain1984fairness} as a measure in the reward function to mitigate the potential unfairness. The Jain's fairness indicator was initially proposed to evaluate the network performance thus it is a favorable choice for our model. By introducing the fairness factor to weigh the reward from each node, balance among nodes can secure.

\begin{figure}
\centerline{\includegraphics{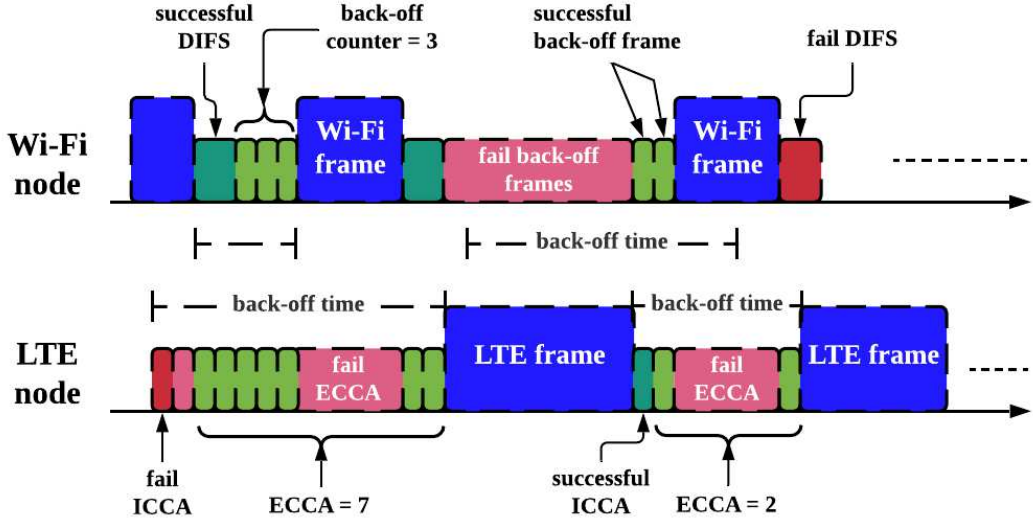}}
	\caption[LTE-LAA spectrum sharing]{Illustration of the LBT-based spectrum sharing mechanism for LTE-LAA and Wi-Fi nodes.}
	\label{fig:lte-wifi}
\end{figure}

As we mentioned in \ref{sec: related_work}, the Wi-Fi standard has been utilizing CSMA/CA for coexistence in the unlicensed spectrum. The CSMA/CA adopts sensing before transmission to minimize packet collision. Before transmission starts, an initial channel sensing for a distributed inter-frame spacing (DIFS) duration is mandatory for each node to evaluate the channel dynamics, access is suppressed if the channel is judged as busy. After channel is judged as idle for initial sensing, the Wi-Fi node then performs an additional back-off sensing to further inspect the channel. For back-off sensing phase, a positive integer will be drawn randomly from a predefined range $[0, CW]$ as a down counter, where $CW$ means contention window. The counter counts down by $1$ every time the channel is sensed idle for a fixed-length time slot. The countdown will freeze for any non-idle result and resume once the it is idle again. Transmission is able to commence after the counter reaches $0$. Stochastic back-off counters drawn by different Wi-Fi nodes avoid collisions by staggering their access timing. Similar to W-Fi, the LTE-LAA standard enables LTE nodes to coexist with other nodes in unlicensed spectrum by implementing LBT mechanism. The main differences with CSMA/CA lie on the duration of initial and back-off sensing slots, and back-off sensing is not mandatory for LTE nodes. According to \cite{3gpp.36.213}, the set of $CW$ values and maximum allowed channel occupation time a LTE node can utilize depend on the channel access priorities. With larger $CW$ value, the LTE nodes are allowed to transmit more sub-frames in one transmission, so there is a trade-off between sensing duration and transmission payload. The initial and back-off channel sensing mechanisms in LTE standard are termed as initial clear carrier assessment (ICCA) and extended clear carrier assessment (ECCA). For simplification, in our scenario the access priorities are equal for both Wi-Fi and LTE nodes, and the back-off sensing shall be performed compulsorily no matter the initial sensing results. A simple example of our spectrum coexistence scenario is illustrated in \ref{fig:lte-wifi}.

\subsection{Model Formulation}

Here we define the components of Dec-POMDP model in our spectrum coexistence scenario.
\begin{itemize}[leftmargin=*]
\item {\em Agents:} each agent is either a LTE-LAA eNB or Wi-Fi AP. There are total $L$ number of LTE-LAA agents and $W$ number of Wi-Fi agents attempting to access the spectrum. Due to limited resource, only a subset of $N=L+W$ agents is able to win the resource at a time. We utilize notation $n$ as agent $n$'s index.
\item {\em Actions:} for each agent, each element $a_i$ in its action set $\{a_1,a_2,...\}$ is a value representing the contention window $CW$. Every time an agent has selected an action $a_i$ from its action set, it will then draw an integer randomly from the interval $\left[0, a_i\right]$ for its back-off counter. All agents share the same set of contention windows while the channel occupation time for LTE agents depends on the selected $CW$ value. Multi-channel operation is not considered, which means all agents access the same channel.
\item {\em States:} each global state $s_k$ represents one spectrum configuration. A configuration is an integer value indicating the amount of agents currently occupying the spectrum. There are total $(N+1)$ number of unique states in the set of global state.
\item {\em Observations:} LTE-LAA and Wi-Fi standards demand agents to inspect the channel activity for additional back-off sensing before transmission. The observation received by agent after each action is defined as the duration from initial sensing starts to the end of back-off sensing, which is the time an agent actually spends in waiting for the channel resource, reflecting the occupation of the channel.
\item {\em Reward Function:} from the aspect of spectrum usage, it is desirable to exploit the channel resource as more efficient as possible. Thus the local reward for agent $n$ at step $t$ is a function of the effective throughput $Th_n^t$ reweighted by the Jain's fairness indicator. It is a cumulative function which depends on current and accumulation of past rewards to reflect the influence of past actions and observations. Denote $PL_n^t$ as the effective transmitted payload without collision, and $D_n^t$ as the duration from initial sensing starts to transmission ends, the global reward received by the nodes which complete their actions at time $t$ is designed as
\end{itemize}

\begin{equation}
\begin{aligned}
    & \text{Global reward }R_t = \sum_{n=1}^{N}r_t^n \\
    & \text{Local cumulative reward }r_t^n = r_{t-1}^n + R_n(t) \\
    & R_n(t) = \ln\left\{\left|\widetilde{Th}_{n}^t\right|+1\right\} \\
    & \widetilde{Th}_n^t = J_n^t Th_n^t, \ \ Th_n^t=\frac{PL_n^t}{D_n^t} \\
\end{aligned}
\label{equ: rwd}
\end{equation}
where $J_n^t$ is the Jain's fairness indicator \cite{jain1984fairness} and is computed by
\begin{equation}
J_n^t = \frac{\left(\sum_{\forall i\neq n}x_i^{t-1} + x_n^t\right)^2}{N\left(\sum_{\forall i\neq n}{x_i^{t-1}}^2 + {x_n^t}^2\right)}, \ \ x_i^t = \frac{Th_i^t}{O_i}, \ \forall i\in [1,N]
\end{equation}
$O_i$ is the theoretical fair throughput for agent $i$; in our algorithm, it is defined as
\begin{equation*}
O_i=\frac{(\text{Maximum data rate})}{(\text{Total spectrum users})}, \ \ \forall i\in[1,N]
\end{equation*}
We want to point put that our Dec-POMDP model is infinite-horizon, which means theoretically the agent-model interaction will never stop (although the agents will stop at some point in practical learning process). This model does not designate an objective state, that is, there is no such state to flag the termination of each episode when some agents have arrived at the state.

\subsection{Policy Representation}

\begin{figure}
\centerline{\includegraphics{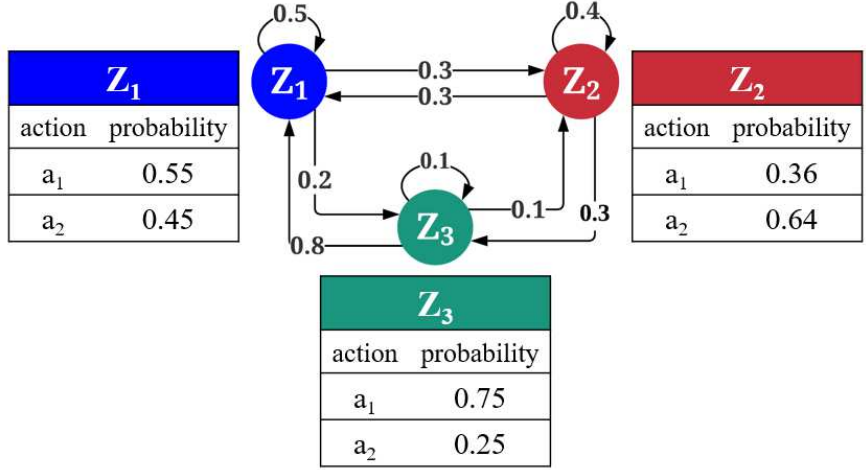}}
	\caption[FSC policy]{Diagram for a simple FSC policy representation with $|\mathcal{Z}|=3$ and $|\mathcal{A}|=2$. Each arch represents the transition from one node to another with some probability; at each node there is a probability distribution over actions.}
	\label{fig:FSC}
\end{figure}

Finite state controller (FSC) is an appropriate policy representation for infinite-horizon Dec-POMDP models \cite{HansenFSC,AmatoFSC} with discrete space. It is subsumed a special case of the regionalized policy representation \cite{li09multi} when each belief region concentrates to one node. In \cite{li09multi,liu_thesis}, each node in the FSC policy is referred as decision state or local belief state and is treated as latent variable, which is integrated out to yield direct mapping from past history of actions and observations to distribution over current actions, thus estimation of the true states can be omitted in learning process. Each node serves as sufficient statistics of histories of past actions and observations, saving memory space from storing histories, thus FSC is efficient in operating on small devices. \ref{fig:FSC} illustrates a simple FSC policy with 3 nodes and 2 actions at each node. The FSC policy representation for agent $n$ can be described by a tuple $\langle\mathcal{A}_n,\mathcal{O}_n,\mathcal{Z}_n,\eta_n,\omega_n,\pi_n\rangle$. $\mathcal{A}_n$ and $\mathcal{O}_n$ are the same in \ref{sec: decPOMDP}; $\mathcal{Z}_n$ is a finite set of nodes; $\eta_n$ is the node distribution at step $t=0$. $\pi_n$ represents the action selection probability at nodes. $\omega_n: \mathcal{Z}_n\times\mathcal{A}_n\times\mathcal{O}_n\rightarrow [0,1]$ expresses the node transition probability, mapping from node, action, and observation sets to node set, indicating how the agent will travel around nodes after an action is performed and an observation is received.
FSC representation fits our problem because of its simplicity and compactness even with large problem space. Since our problem does not define an explicit end point, map-like policy representations will not work properly. Even though the observation or reward space is enormous, practically there is only a comparatively small part of it assigned positive rewards, which is desired for the agent. The cyclic graph of FSC representation captures the essential parts of our infinite-horizon POMDP model and yields a (ideally) compact structure for the optimal policy, which makes FSC policy popular in various reinforcement learning problems.

\subsection{Nonparametric Policy Prior}

One of the main problems in learning the FSC policies for decentralized agents is determining the sizes of the policies, i.e, number of nodes. As we have emphasized, our coexistence problem is dynamic and non-cooperative. With decentralized learning, the local action and observation sets possessed by each agent differ, causing the node sets and transitions between nodes in different policies deviate from each other. Parametric models impose strong assumption over policy structures, yielding fixed-size policies, which is not favorable in decentralized cases. Bounding the space of policy representation may converge the learning process to sub-optimal results. Meanwhile, nonparametric model treats the FSC size as additional variable, which enables it to accommodate unbounded possibilities of policy structures and transition probabilities, allowing each agent to achieve its optimal result, respectively.
\begin{defn}
Providing the FSC policy representation, the stick-breaking process is utilized to generate the prior distributions for node transition probabilities $\omega_n$, and Dirichlet distribution is adopted for prior distribution over actions $\pi_n$ at each node. To accommodate more flexibility in stick-breaking prior, Gamma distribution is placed over $\alpha$ in beta distribution as hierarchical prior \textup{\cite{LiuSBPR}}:
\[
\begin{aligned}
    & \eta_n^1=u_n^1,\ \eta_n^i=u_n^i\prod_{m=1}^{i-1}\left(1-u_n^m\right) \\
    & \omega_{n,a,o}^{i,1}=V_{n,a,o}^{i,1},\  \omega_{n,a,o}^{i,1:j}=V_{n,a,o}^{i,1:j}\prod_{m=1}^{j-1}\left(1-V_{n,a,o}^{i,m}\right) \\
    & u_n^{1:\infty}\sim\textup{Beta}(1,\rho_{n}), \ \ \rho_{1:N}\sim\textup{Gamma}(e,f) \\ 
    & V_{n,a,o}^{i,1:\infty}\sim\textup{Beta}(1,\alpha_{n,a,o}^{i}), \ \ \alpha_{n,a,o}^{1:\infty}\sim\textup{Gamma}(c_{n,a,o},d_{n,a,o}) \\
    & \pi_{n,i}^{1:|\mathcal{A}_n|}\sim\textup{Dirichlet}\left(\theta_{n,i}^{1:|\mathcal{A}_n|}\right)
\end{aligned}
\]
for node indices $i,j=1,...,\infty$
\label{def: SBPR}
\end{defn}
Hyper-parameters $(c, d, e, f, \theta)$ determine the distributions of $\eta$, $\omega$, and $\pi$. $|\cdot|$ means the cardinality of set. For notational elegance, we utilize the same abbreviation in \cite{LiuSBPR}. Continuous sequence $(i,i+1,...,j)$ is abbreviated to $i:j$. For example, $\omega_{n,a,o}^{i,1:j}$ compresses $(\omega_{n,a,o}^{i,1},...,\omega_{n,a,o}^{i,j})$, representing agent $n$'s node transition probabilities to nodes $1,...,j$ after performing $a$ at node $i$ and observing $o$. Similarly, $\left(\pi_{n,i}^{1},...,\pi_{n,i}^{|\mathcal{A}_n|}\right)=\pi_{n,i}^{1:|\mathcal{A}_n|}$ means the probabilities of selecting actions $a_1,...,a_{|\mathcal{A}_n|}$ at node $i$.

\subsection{Global Empirical Value Function}

\begin{figure}
\centerline{\includegraphics[width=0.9\linewidth]{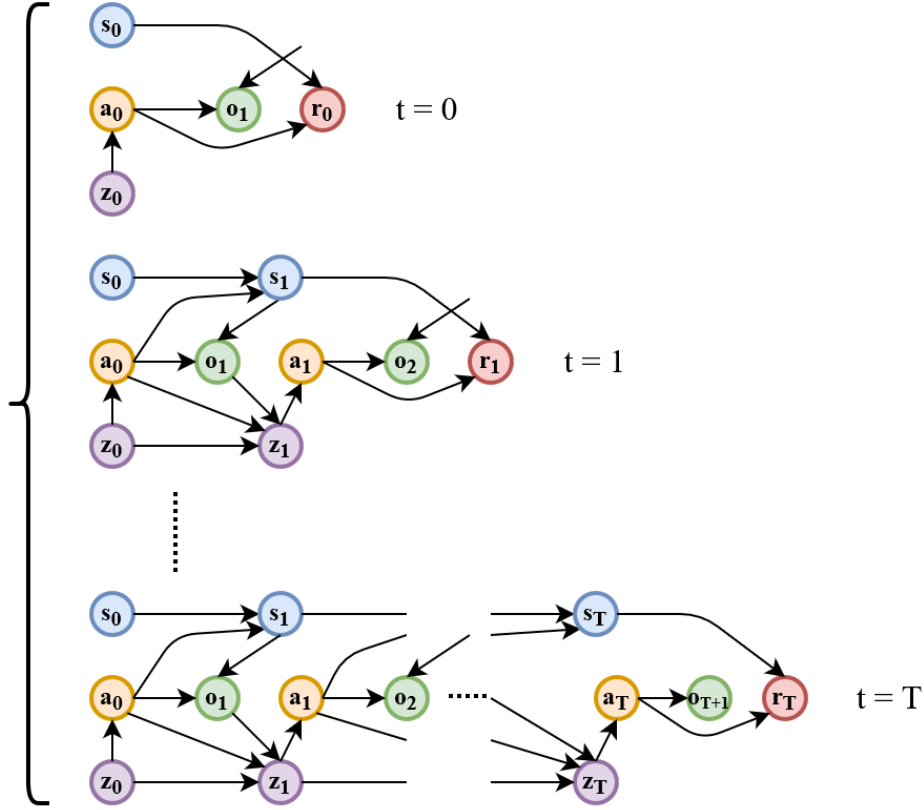}}
	\caption[Mixture of DBNs]{Decomposing the DBN of our POMDP model into mixture of sub-networks, where reward only emits at the end of each sub-network.}
	\label{fig:DBN}
\end{figure}

In general, the objective of reinforcement learning is to maximize the overall value. In order to adopt Bayesian approach, the value function is transformed into likelihood to reflect the value from dataset given policy \cite{toussaintDBN}. A Dec-POMDP can be formulated as one single dynamic Bayes network (DBN) with a binary reward variable $R$ at each time step. This DBN can be further decomposed into an infinite mixture of DBNs \cite{kumarDBN}, where reward only emerges at the end of each DBN. \ref{fig:DBN} illustrates the result of decomposing the Bayes network of our POMDP model into mixture of sub-networks. There is only one unique DBN for sub-network with number of steps $T=t$. Denote $r_T(\Theta)$ as immediate reward received with policy $\Theta$ in $T$-th sub-networks, the value $\hat{r}_T(\Theta)$ obtained by by normalizing $r_T(\Theta)$ into range $[0,1]$ is proportional to the likelihood $p(R=1|T,\Theta)$,
\begin{equation}
    \hat{r}_T(\Theta)=\frac{r_T(\Theta)-R_{\text{min}}}{R_{\text{max}}-R_{\text{min}}}\propto p(R=1|T,\Theta)
\label{equ: norm_rwd}
\end{equation}
This implies maximizing likelihood $p(R=1|T,\Theta)$ is equivalent to maximizing the normalized reward. After placing geometric distribution over the mixture sub-networks with parameter equal to the discount factor $\gamma$, the joint likelihood $p(R=1|\Theta)$ is obtained by marginalizing $T$,
\begin{equation}
\begin{aligned}
    p(R=1|\Theta) &=\sum_{t=0}^{T}p(t)p(R=1|t,\Theta) \\
    &=\sum_{t=0}^{T}(1-\gamma)\gamma^t p(R=1|t,\Theta) \\
    &=\sum_{t=0}^{T}(1-\gamma)\gamma^t \frac{r_t(\Theta)-R_{\text{min}}}{R_{\text{max}}-R_{\text{min}}} \\
    &=\frac{1-\gamma}{R_{\text{max}}-R_{\text{min}}}\left[\sum_{t=0}^{T}\gamma^t r_t(\Theta)-\sum_{t=0}^{T}\gamma^tR_{\text{min}}\right] \\
    &=\frac{1-\gamma}{R_{\text{max}}-R_{\text{min}}}\hat{V}(\Theta),
\end{aligned}
\label{equ: likelihood}
\end{equation}
where $\hat{V}(\Theta)$ is the shifted value function for policy $\Theta$. Maximizing this likelihood amounts to maximizing the value of policy $\Theta$. In \cite{li09multi} Li proposed an empirical value function $\hat{V}(\mathcal{D}^K;\Theta)$ to acquire the value of desired policy $\Theta$ with $K$ trajectories. 
\begin{defn}
Define $k$-th history of agent $n$ from time $0$ to $t$ as the sequence $(a_{n,0}^k,...,a_{n,t-1}^k;o_{n,1}^k,...,o_{n,t}^k)=(a_{n,0:t-1}^k,o_{n,1:t}^k)=h_{n,t}^k$, and $k$-th trajectory $\mathcal{D}^k$ of length $T_k$ as the sequence $(\vec{a}_0^k,r_0^k,\vec{o}_1^k,...,\vec{o}_{T_k}^k,\vec{a}_{T_k}^k,r_{T_k}^k)$. The value $\hat{V}(\mathcal{D}^K;\Theta)$ is the expected value of discount sum of rewards with respect to reweighted policy:
\[
\begin{aligned}
    & \hat{V}(\mathcal{D}^K;\Theta) \\
    &=\textup{E}_{\Theta}\left[\sum_{k=1}^{K}\sum_{t=0}^{T_k}\gamma^t\frac{(r_t^k-R_{\textup{min}})}{K}\right] \\
    &=\sum_{k=1}^{K}\sum_{t=0}^{T_k}\frac{\prod_{\tau=0}^{t}\prod_{n=1}^{N}p(a_{n,\tau}^k|h_{n,\tau}^k,\Theta)}{\prod_{\tau=0}^{t}\prod_{n=1}^{N}p(a_{n,\tau}^k|h_{n,\tau}^k,\Pi)}\gamma^t\frac{(r_t^k-R_{\textup{min}})}{K}
\end{aligned}
\]
\label{def: emp_likelihood}
\end{defn}
$\prod_{\tau=0}^{t}\prod_{n=1}^{N}p(a_{n,\tau}^k|h_{n,\tau}^k,\Theta)$ can be substituted with $p(\vec{a}_{0:t}^{k},\vec{z}_{0:t}^{k}|\vec{o}_{1:t}^{k},\Theta)$ (proof in \ref{ch: em_value}). $\Theta$ is reweighted by the behavior policy $\Pi$ which is utilized for collecting trajectories. By law of large number, $\hat{V}(\mathcal{D}^K;\Theta)$ approximates $\hat{V}(\Theta)$ as number of trajectories $K$ approaches infinity. \ref{def: emp_likelihood} enables us to incorporate trajectories from other sources to compute the likelihood rather than collecting them on our own. By combining equation \ref{equ: likelihood} and \ref{def: emp_likelihood}, the likelihood is connected to the empirical value function,
\begin{equation*}
    p(\mathcal{D}^K|\Theta)\propto p(R=1|\Theta)\propto \hat{V}(\mathcal{D}^K;\Theta)
\end{equation*}

\subsection{Posterior Inference}

Providing prior distributions and likelihood function, the objective is to infer the posterior distribution $p(\Theta|\mathcal{D}^K)$. Markov chain Monte-Carlo method can approximate the exact posterior distributions by directly sampling them, however, its computationally demanding nature refrains it from scaling to complex problems. In contrast, variational inference (VI) provides an feasible alternative to infer the optimal results of the approximate posterior distributions. By casting the inference of distributions as an optimization problem, it is convenient to apply optimization techniques to accelerate the computations, and thus is more efficient for complex problems. Its results are guaranteed to be the optimal possible approximate to the objective distributions in the distribution family. To achieve VI results, we derive the expectation term for joint likelihood
\begin{equation}
\medmuskip=1mu
\thinmuskip=1mu
\thickmuskip=1mu
\begin{aligned}
&\text{E}_q\left[\ln \hat{V}(\mathcal{D}^K;\Theta)p(\Theta)p(\rho)p(\alpha)\right]\\
&=\text{E}_q\left[\ln \sum_{k=1}^{K}\sum_{t=0}^{T_k}\frac{\prod_{\tau=0}^{t}\prod_{n=1}^{N}p(a_{n,\tau}^k|h_{n,\tau}^k,\Theta)}{\prod_{\tau=0}^{t}\prod_{n=1}^{N}p(a_{n,\tau}^k|h_{n,\tau}^k,\Pi)}\gamma^t\frac{(r_t^k-R_{\textup{min}})}{K}p(\Theta)p(\rho)p(\alpha)\right] \\
&=\text{E}_q\left[\ln \sum_{k=1}^{K}\frac{1}{K}\sum_{t=0}^{T_k}\Tilde{r}_t^k p(\vec{a}_{0:t}^{k}|\vec{o}_{1:t}^{k},\Theta)p(\Theta)p(\rho)p(\alpha)\right]\\
&=\text{E}_q\left[\sum_{k=1}^{K}\frac{1}{K}\sum_{t=0}^{T_k}\sum_{\vec{z}_{0:t}^k=1}^{|Z|}\ln \left[\Tilde{r}_t^k p(\vec{a}_{0:t}^{k},\vec{z}_{0:t}^{k}|\vec{o}_{1:t}^{k},\Theta)p(\Theta)p(\rho)p(\alpha)\right]\right]\\
&=\sum_{k=1}^{K}\frac{1}{K}\sum_{t=0}^{T_k}\sum_{\vec{z}_{0:t}^k=1}^{|Z|}\int q(\Theta)q(\rho)q(\alpha)q(\vec{z}_{0:t}^k)\ln \Tilde{r}_t^k p(\vec{a}_{0:t}^{k},\vec{z}_{0:t}^{k}|\vec{o}_{1:t}^{k},\Theta)d\Theta d\rho d\alpha \\
&+\sum_{k=1}^{K}\frac{1}{K}\sum_{t=0}^{T_k}\sum_{\vec{z}_{0:t}^k=1}^{|Z|}\text{E}_q\left[\ln p(\Theta)+\ln p(\rho)+\ln p(\alpha)\right] \\
&=\sum_{k=1}^{K}\frac{1}{K}\sum_{t=0}^{T_k}\sum_{\vec{z}_{0:t}^k=1}^{|Z|}\int q(\Theta)q(\vec{z}_{0:t}^k)\ln \Tilde{r}_t^k p(\vec{a}_{0:t}^{k},\vec{z}_{0:t}^{k}|\vec{o}_{1:t}^{k},\Theta)d\Theta \\
&+\text{E}_q\left[\ln p(\Theta)\right]+\text{E}_q\left[\ln p(\rho)\right]+\text{E}_q\left[\ln p(\alpha)\right] \\
&=\text{E}_{q(\Theta,z)}\left[\ln \Tilde{r}_t^k p(\vec{a}_{0:t}^{k},\vec{z}_{0:t}^{k}|\vec{o}_{1:t}^{k},\Theta)\right]+\text{E}_{q(\Theta,\rho,\alpha)}\left[\ln p(\Theta)\right] \\
&+\text{E}_{q(\rho)}\left[\ln p(\rho)\right]+\text{E}_{q(\alpha)}\left[\ln p(\alpha)\right]
\end{aligned}
\label{equ: ELBO_p}
\end{equation}
where $\Tilde{r}_t^k=\gamma^t\frac{r_t^k-R_{\text{min}}}{\prod_{n=1}^{N}p(a_{n,0:t}^k|o_{n,1:t}^k,\Pi)}$. $\Theta$ denotes the policy parameters $(u, V, \pi)$. The probability of node transition history $p(z_{n,0:t}^k|a_{n,1:t}^k,o_{n,1:t}^k,\Theta)$ will also be inferred since $(\eta, \omega, \pi)$ are conditional on $z$. By assuming mean-field factorization over parameters, the expectation over $q$ distribution can be derived
\begin{equation}
    \begin{aligned}
        &\text{E}_q\left[\ln q(\Theta, \rho,\alpha)q(\vec{z}_{0:t}^k)\right]\\
        &=\text{E}_q\left[\ln q(\Theta)q(\rho)q(\alpha)q(\vec{z}_{0:t}^k)\right]\\
        &=\text{E}_q\left[\ln q(\Theta)\right]+\text{E}_q\left[\ln q(\rho)\right]+\text{E}_q\left[\ln q(\alpha)\right]+\text{E}_q\left[\ln q(\vec{z}_{0:t}^k)\right]\\
        &=\text{E}_{q(\Theta)}\left[\ln q(\Theta)\right]+\text{E}_{q(\rho)}\left[\ln q(\rho)\right]+\text{E}_{q(\alpha)}\left[\ln q(\alpha)\right]+\text{E}_{q(z)}\left[\ln \prod_{n=1}^{N}q(z_{n,0:t}^k)\right]
    \end{aligned}
\label{equ: ELBO_q}
\end{equation}
Combining \ref{equ: ELBO_p} and \ref{equ: ELBO_q}, $\textup{ELBO}(q)$ is expressed as
\begin{equation}
    \textup{ELBO}(q)=\text{E}_q\left[\ln \hat{V}(\mathcal{D}^K;\Theta)p(\Theta)p(\rho)p(\alpha)\right]-\text{E}_q\left[\ln q(\Theta,\rho,\alpha)q(\vec{z}_{0:t}^k)\right]
\label{equ: ELBO_final}
\end{equation}
Mean-field assumption factorizes the joint variational $q(\Theta)$ into product of marginal distributions $q(u)q(V)q(\pi)$ conditional on their corresponding parameters. Since the likelihood is discrete distribution, the Dirichlet distribution and Dirichlet process for policy priors are conjugate prior. The true posterior distributions for $(u,V,\pi,\rho,\alpha)$ can reasonably be assumed to belong to the same family of their corresponding prior distributions. Thus the variational distributions $q$ for posterior inference are defined in the form:
\begin{equation}
    \begin{aligned}
    &q(z_{n,0:t}^{k})=\Tilde{\nu}_t^k p(z_{n,0:t}^k|a_{n,0:t}^k,o_{n,1:t}^k,\Tilde{\Theta}), \  \forall (n,k,t) \text{ indices} \\
    &q(u_n^i)=\text{Beta}(\delta_n^i,\mu_n^i), \, \forall (n,i) \text{ indices} \\
    &q(V_{n,a,o}^{i,j})=\text{Beta}(\sigma_{n,a,o}^{i,j},\lambda_{n,a,o}^{i,j}), \  \forall (n,a,o,i,j) \text{ indices} \\
    &q(\rho_n)=\text{Gamma}(g_n,h_n), \  \forall n \text{ indices} \\
    &q(\alpha_{n,a,o}^{i})=\text{Gamma}(a_{n,a,o}^{i},b_{n,a,o}^{i}), \  \forall (n,a,o,i) \text{ indices} \\
    &q(\pi_{n,i})=\text{Dirichlet}\left(\phi_{n,i}^{1},...,\phi_{n,i}^{|\mathcal{A}_n|}\right), \  \forall (n,i) \text{ indices} \\
    &\Tilde{\nu}_t^k =\gamma^t (r_t^k-R_{\text{min}}) \frac{\prod_{n=1}^{N}p(a_{n,0:t}^{k}|o_{n,1:t}^{k}, \Tilde{\Theta})}{\prod_{n=1}^{N}p(a_{n,0:t}^{k}|o_{n,1:t}^{k}, \Pi)\hat{V}(\mathcal{D}^K;\Tilde{\Theta})} 
\end{aligned}
\label{equ: q_dist}
\end{equation}
$\Tilde{\Theta}=(\Tilde{\eta},\Tilde{\pi},\Tilde{\omega})$ is the point estimate of optimal policy parameters in variational inference at previous iteration. It is worth to note that each node transition probability $q(z_{n,t}^{k})$ is a multinomial distribution. For simplicity, mean-field variational distribution $q_{n,t}^k(z_{n,0:t}^{k})$ is estimated using expectation maximization \cite{li09multi}, although Bayesian inference can also be adopted by placing another Dirichlet process prior over it. By taking partial derivative on $\text{ELBO}(q)$ with respect to each $q$ distribution and setting it to zero, the coordinate ascent VI (CAVI) is formulated for each optimal variational distribution $q^*$.
\begin{theorem}
With conjugate prior and mean-field assumption, the computation of each variational distribution $q$ can be simplified to its parameter update in \textup{\ref{equ: q_dist}}:
\[
\medmuskip=1mu
\thinmuskip=1mu
\thickmuskip=1mu
\begin{aligned}
    &\delta_n^i=1+\sum_{k=1}^{K}\frac{1}{K}\sum_{t=0}^{T_k}q_{n,t}^k(z_{n,0}^{k}=i) \\
    &\mu_n^i=\frac{g_n}{h_n}+\sum_{k=1}^{K}\frac{1}{K}\sum_{t=0}^{T_k}\sum_{m=i+1}^{|\mathcal{Z}_n|}q_{n,t}^k(z_{n,0}^{k}=m) \\
    &\phi_{n,i}^{a}=\theta_{n,i}^{a}+\sum_{k=1}^{K}\frac{1}{K}\sum_{t=0}^{T_k}\sum_{\tau=0}^{t}q_{n,t}^k(z_{n,\tau}^{k}=i)\mathbb{I}(a_{n,\tau}^{k}=a) \\
    &\sigma_{n,a,o}^{i,j}=1+\sum_{k=1}^{K}\frac{1}{K}\sum_{t=0}^{T_k}\sum_{\tau=1}^{t}q_{n,t}^k(z_{n,\tau-1}^{k}=i,z_{n,\tau}^{k}=j)\mathbb{I}(a_{n,\tau-1}^{k}=a,o_{n,\tau}^{k}=o) \\
    &\lambda_{n,a,o}^{i,j}=\frac{a_{n,a,o}^{i}}{b_{n,a,o}^{i}}+\sum_{k=1}^{K}\frac{1}{K}\sum_{t=0}^{T_k}\sum_{\tau=1}^{t}\sum_{m=j+1}^{|\mathcal{Z}_n|}q_{n,t}^k(z_{n,\tau-1}^{k}=i,z_{n,\tau}^{k}=m)\mathbb{I}(a_{n,\tau-1}^{k}=a,o_{n,\tau}^{k}=o) \\
    &g_n=e+|\mathcal{Z}_n|, \ \ h_n=f-\sum_{i=1}^{|\mathcal{Z}_n|}\left[\Psi(\mu_{n}^{i})-\Psi(\delta_{n}^{i}+\mu_{n}^{i})\right] \\
    &a_{n,a,o}^{i}=c_{n,a,o}+|\mathcal{Z}_n|, \ \ b_{n,a,o}^{i}=d_{n,a,o}-\sum_{j=1}^{|\mathcal{Z}_n|}\left[\Psi(\lambda_{n,a,o}^{i,j})-\Psi(\sigma_{n,a,o}^{i,j}+\lambda_{n,a,o}^{i,j})\right]
\end{aligned}
\]
where
\[
\begin{aligned}
    & q_{n,t}^k(z_{n,\tau}^{k}=i)=\Tilde{\nu}_t^k p(z_{n,\tau}^{k}=i\vert a_{n,0:t}^k,o_{n,1:t}^k,\Tilde{\Theta}) \\
    & q_{n,t}^k(z_{n,\tau-1}^{k}=i,z_{n,\tau}^{k}=j)=\Tilde{\nu}_t^k p(z_{n,\tau-1}^{k}=i,z_{n,\tau}^{k}=j\vert a_{n,0:t}^k,o_{n,1:t}^k,\Tilde{\Theta})
\end{aligned}
\]
are marginal distributions of $q_{n,t}^k(z_{n,0:t}^{k})$ for $\tau=0,...,t$.
\label{thm: q_update}
\end{theorem}

$\Psi(\cdot)$ is the digamma function. The detailed derivations of \ref{thm: q_update} are elaborated in \ref{ch: VI_update}. Each Bayesian learning iteration for our Dec-POMDP model is demonstrated as the following

\begin{algorithm}
\SetKwInOut{Input}{Input}\SetKwInOut{Output}{Output}\SetKwInOut{Initialize}{Initialize}\SetKw{Return}{Return}
\SetKwIF{If}{ElseIf}{Else}{if}{}{}{}{end}
\SetAlgoLined
\Input{$p(\Theta_n)$, $p(\rho_n)$, $p(\alpha_n)$, trajectories $\mathcal{D}^k$, $k=1,...,K$}
\Initialize{initial $\textsc{ELBO}_0(q)$}
 \For{ $\textup{Iter}=1,...,\textup{max}$}{
  Update $\Tilde{\Theta}_n=(\Tilde{\eta}_n,\Tilde{\omega}_n,\Tilde{\pi}_n)$ for $n=1,...,N$ \\
  Compute each marginal $q_{n,t}^k(z_{n,\tau}^k)$ for $\tau=0,...,T_k$\\
  Compute each $q^*(\Theta_n)$, $q^*(\rho_n)$, and $q^*(\alpha_n)$ according to \ref{thm: q_update}\\
  Compute $\textsc{ELBO}_{\textup{Iter}}(q)$ by \ref{equ: ELBO_final} \\
  $\Delta\textsc{LB}(q)=\vert(\textsc{ELBO}_{\textup{Iter}}(q)-\textsc{ELBO}_{\textup{Iter}-1}(q))/\textsc{ELBO}_{\textup{Iter}-1}(q)\vert$ \\
  \If{$\Delta\textsc{LB}(q) < 10^{-5}$}{
   break\;
   }
 }
\Output{variational distributions $q^*(\Theta_n)$, $q^*(\rho_n)$, and $q^*(\alpha_n)$ for $n=1,...,N$}
\caption{CAVI for Bayesian Reinforcement Learning}
\label{alg: learning}
\end{algorithm}
\section{Performance Evaluation}
\label{sec: simulation}

In this section we detail the system setup for performance evaluation and demonstrate the simulation results along with discussions.
%
\begin{table}[htbp]
\begin{center}
\begin{tabular}{|l|l|}
\hline
\textbf{Parameter Name} & \textbf{Value} \\
\hline \hline 
Number of LTE eNB & 2 \\ 
Number of Wi-Fi AP & 2 \\ 
Number of channel & 1 \\
Channel bandwidth & 20~MHz \\
DIFS duration & 34~$\mu$s \\
Wi-Fi back-off slot & 9~$\mu$s \\
ICCA duration & 43~$\mu$s \\
ECCA slot & 9~$\mu$s \\
Contention window & 15,31,63,127,255,511,1023 \\
LTE sub-frames per transmission & 3,6,8,10~ms \\
Wi-Fi packets per transmission & 1 \\
size of Wi-Fi packet & 15000~bytes \\
Transmission rate & 30~Mbps \\
Discount factor $\gamma$ & 0.9 \\
\hline
\end{tabular}
\end{center}
\caption{Pre-defined parameters}
\label{tab: simu_param}
\vspace{-5mm}
\end{table}

\ref{tab: simu_param} lists the parameters for framing our simulation environment. For simplicity, the sets of contention windows are identical for both LTE and Wi-Fi agents. We simulates the time-domain spectrum coexistence, which means there is only one channel in the spectrum. \cite{3gpp.36.213} defines the maximum channel occupation time and contention windows for LTE agents for different spectrum access priorities. In our scenario, the channel occupation time each LTE agent can utilize depends on the contention window it selects in consideration of fair sharing with Wi-Fi agents. For instance, if a LTE agent selects $CW$ value 15 for its back-off sensing, then it will transmit its data frame for 3ms. This occupation time is 6ms for values $\{31, 63\}$, 8ms for $\{127, 255\}$, and 10ms for $\{511, 1023\}$. On the other hand, Wi-Fi packet is fixed-size no matter which $CW$ value it selects. Each Wi-Fi packet amounts of 15000 bytes including overhead. During back-off sensing, the agent performs spectrum sensing each 1$\mu$s to assess the channel occupation status. However, the correctness of assessment may be affected by path loss, fading, and shadowing effect between the transmitters and receivers. Each agent on the channel has probability $p_e$ to be judged as off-channel in each sensing slot. Each back-off sensing slot is considered as clear if the channel is assessed as occupied no more than 5$\mu$s out of 9$\mu$s. A Wi-Fi packet or LTE sub-frame is assumed to be lost if collision happens when it is being transmitted, and one fail sub-frame will not affect other sub-frames in the same LTE data frame. All wireless agents have infinite amount of data to transfer, which means they request the channel resource permanently. Finally, all values are rounded to integer for discrete model.

For the optimization of \ref{alg: learning}, the FSC policies for all agents are learned from $K=200$ episodes with each episode of length $T=50$. To accelerate the optimization, cross validation was implemented in advance for better configuring the hyper-parameters of the prior distributions. As a result, the hyper-parameters are chosen as $c=e=0.1$, $d=f=100$ in order to encourage smaller FSC policy structures. To initialize the value of the node cardinality for each FSC policy, all episodes collected are converted to FSC structures before learning process by method similar to \cite{AmatoFSCinit}.
\begin{figure}[t]
    \centering
    \begin{subfigure}[b]{0.5\textwidth}
        \centering
        \includegraphics[width=\textwidth]{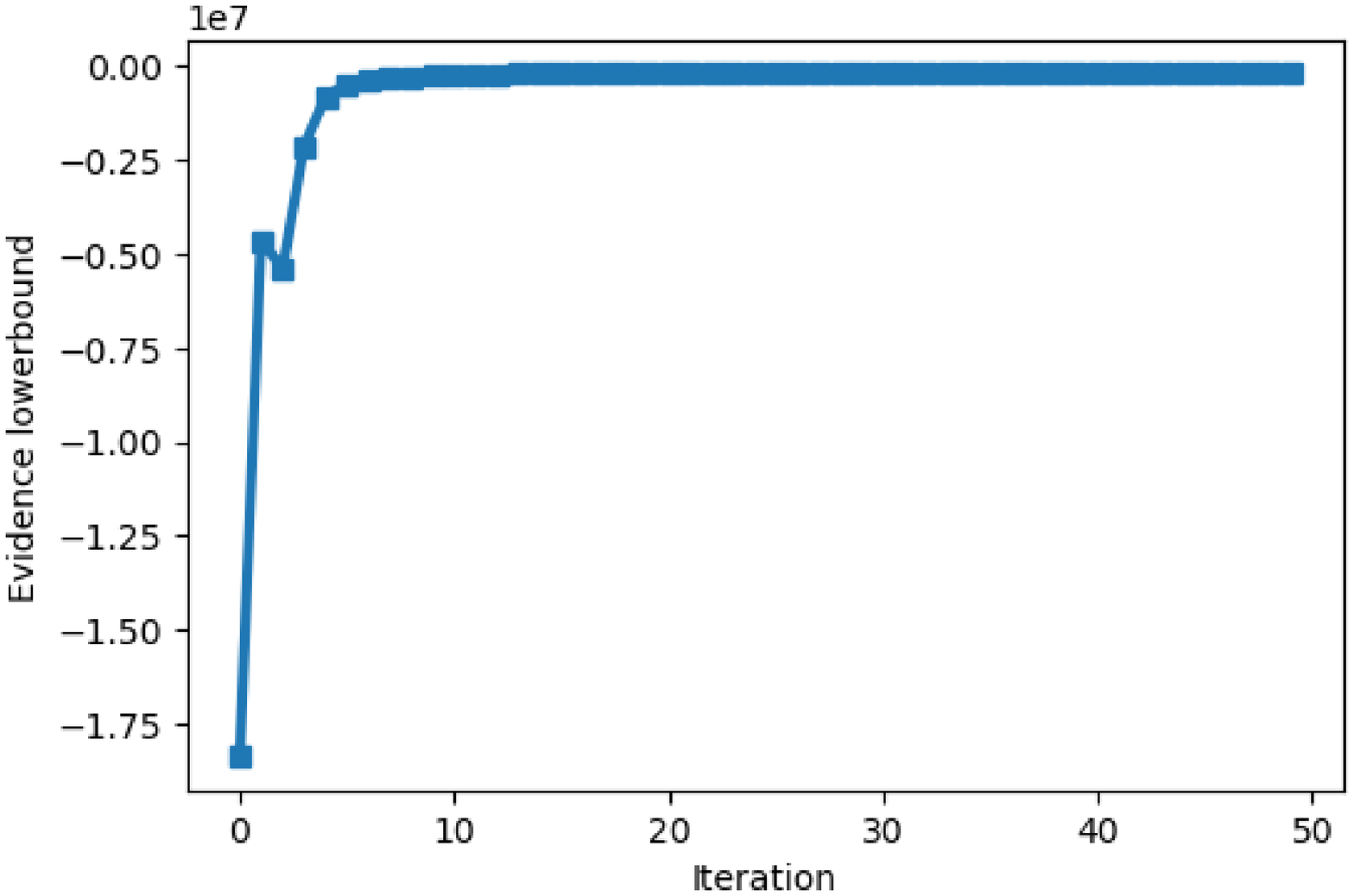}
        \caption[]{{\small ELBO value}}
        \label{fig:ELBO_evolution}
    \end{subfigure}
    \hfill
    \begin{subfigure}[b]{0.477\textwidth}  
        \centering 
        \includegraphics[width=\textwidth]{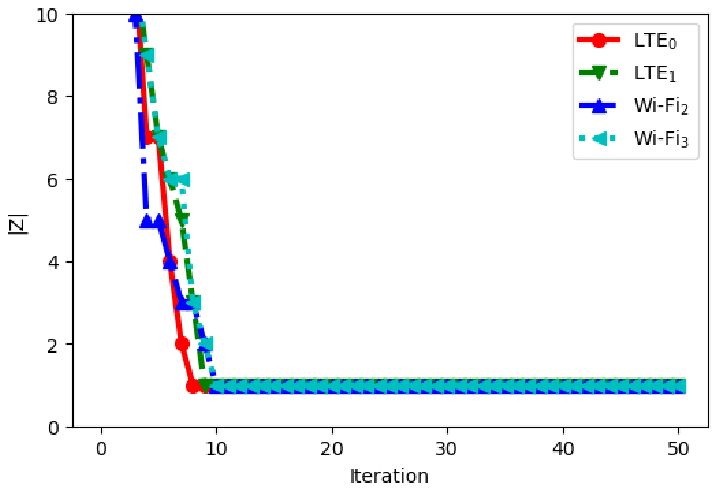}
        \caption[]{{\small Number of FSC nodes}}
        \label{fig:Z_evolution}
    \end{subfigure}
\caption[Evolution of The ELBO Value and Policy Size]{Evolution of the ELBO value and policy size. (a) The convergence of evidence lower bound, (b) The parameter $h$ are fluctuating around a certain level for each agent.}
\label{fig:ELBO_Z_evolution}
\end{figure}
\begin{figure}[t]
\centerline{\includegraphics{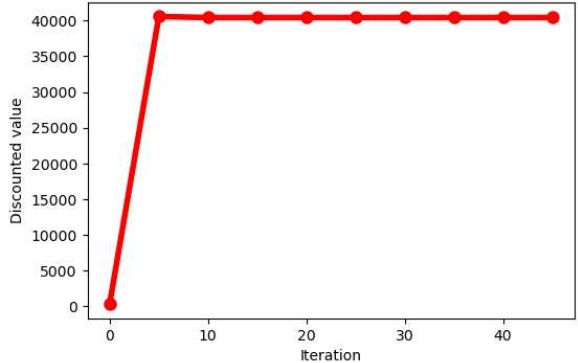}}
	\caption[Discount Value]{Evolution of the discount value of the joint policy.}
	\label{fig:discount_value}
\end{figure}
\begin{figure}[t]
    \centering
    \begin{subfigure}[b]{0.483\textwidth}
        \centering
        \includegraphics[width=\textwidth]{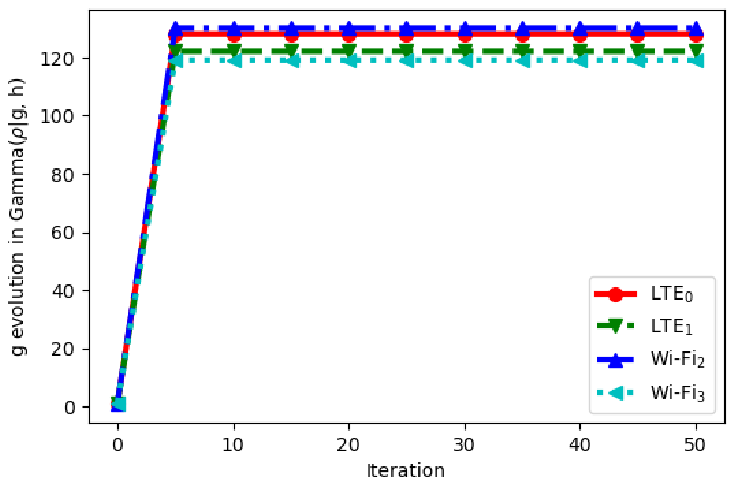}
        \label{fig:g_history}
    \end{subfigure}
    \hfill
    \begin{subfigure}[b]{0.5\textwidth}  
        \centering 
        \includegraphics[width=\textwidth]{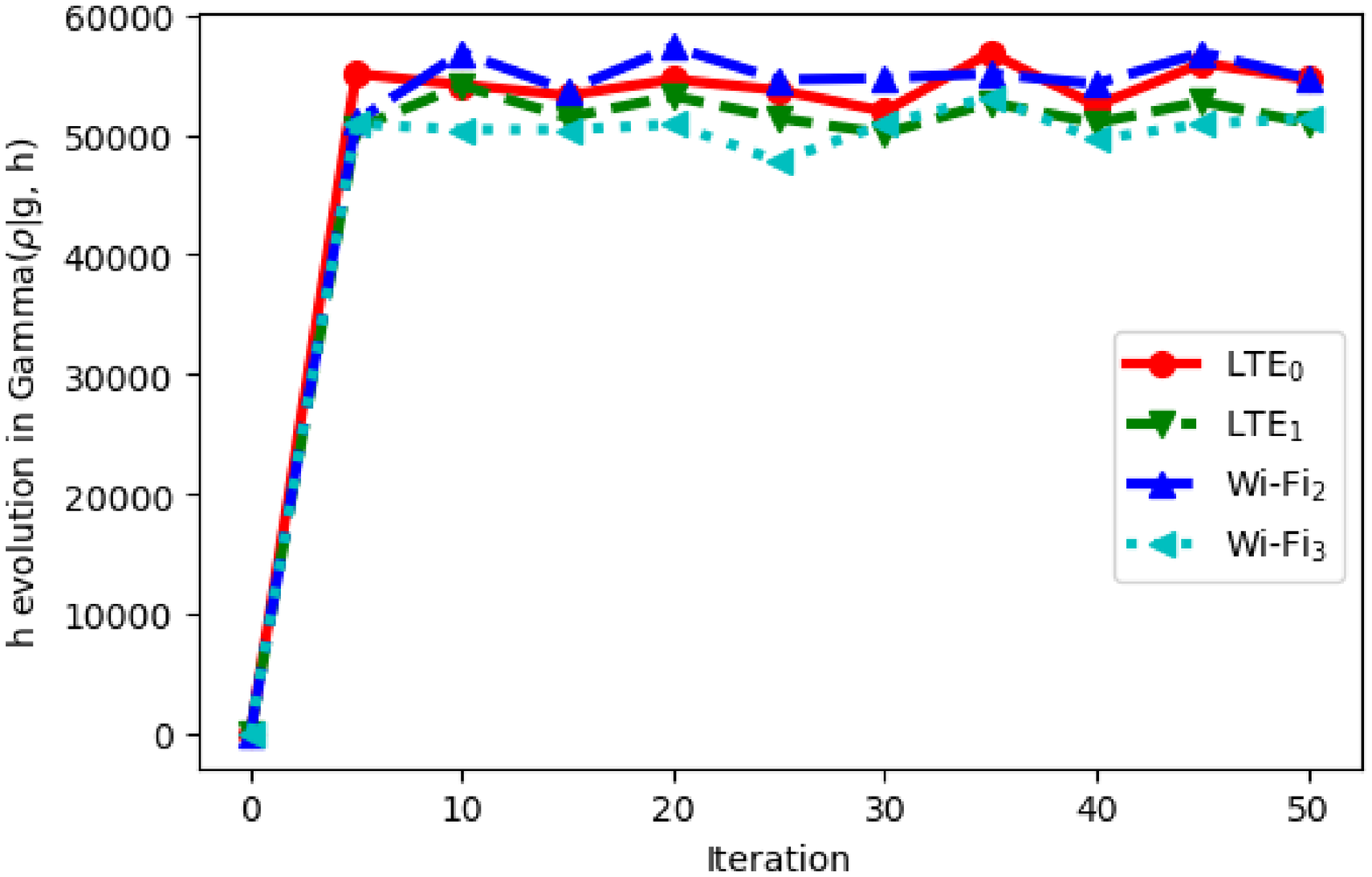}
        \label{fig:h_history}
    \end{subfigure}
\caption[Evolution of The parameters for $q(\rho|g, h)$]{Evolution of the parameters for $q(\rho|g, h)$; the parameter $g$ stays constant during the iterations, while $h$ are fluctuating around a certain level for each agent.}
\label{fig:para_evolution}
\end{figure}
\subsection{Convergence of Variational Inference}
\ref{fig:ELBO_evolution} illustrates the convergence of the evidence lower bound. As iteration was progressing, the lower bound was ascending fast to a certain value and the iteration ended once the value fluctuation was lower than predefined threshold. As lower bound was converging, \ref{fig:Z_evolution} demonstrates how the node cardinalities of FSC policies for all agents gradually optimized to the minimum values required for maximizing the lower bound. It is interesting to indicate that due to the rich-gets-richer property, the sizes of node sets all shrinked to $1$ with maximum lower bound and discounted value, which means all policies reduced to one-state controller similar to the multi-armed bandit. The evolution of the discounted value, which is the sum of discounted rewards of all trajectories, as a function of iteration is illustrated in \ref{fig:discount_value}. Similar to the convergence of lower bound, the ascending discounted value certified the improvement of FSC policies with iterations. The evolution of parameters $(g,h)$ for $q(\rho)$ for each agent is plotted in \ref{fig:para_evolution}; as equations in \ref{thm: q_update}, the posterior of $g$ is a constant for each agent while $h$ fluctuated at different level since it was jointly optimized with other $q$ distributions.
\section{Conclusion}
\label{sec: conclusion}

As the generation of wireless communication advances, coexistence in unlicensed spectrum is becoming an urgent issue calling for solutions. Many solutions have been proposed, however they all leave some common issues unanswered. We loosened assumptions imposed in previous methods and proposed a model close to the real scenario. Meanwhile, reinforcement learning is a thriving topic and has been applied to many classes of problem, including coexistence. Bayesian inference over reinforcement learning provides a handy way to encrypt prior knowledge in learning process so that the need for data is reduced. Nonparametric model over priors enables the learning agents to determine the final representations without being restricted by the inappropriate assumptions and simplifies the learning process while sill achieving excellent result. The incorporation of empirical value function and variational inference transformed the iterative process of dynamic programming into optimization problem, which is superior to conventional reinforcement learning method especially on complex problem.

\subsection{Future Works}
A never-ending question in reinforcement learning is how to balance the exploration and exploitation during learning process. Exploitation enables the agent to exploit the optimal policy learned so far to maximize the value while exploration provides opportunity to search uncharted region of the policy space for potentially better result. Generally, exploration should be encouraged for collecting information in the early phase of learning process. As learning process progresses, the uncertainty becomes less and less, exploitation gradually takes over to settle the learned policy. The trick is how to determine exploration rate with learning iterations. A plummeting curve may incur premature results while flat curve may fail to converge. In this work we utilized $\epsilon$-greedy method to tackle the exploration-exploitation question at each step. $\epsilon$-greedy method determines exploration or exploitation for action selection by its parameter $\epsilon\in (0,1)$. In each learning iteration when the agent is selecting an action, a value $u$ is uniformly drawn in interval $[0,1]$; if $u>\epsilon$, the agent performs exploitation and selects action according to optimal policy learned so far, otherwise the action will be selected uniformly from all actions available for exploration. To demonstrate the trade-off between exploration and exploitation, two $\epsilon$-greedy curves are implemented: both start with exploration rate $0.9$ and end at $0.2$, but have different changing rates.

In addition to those we have accomplished, there are still outstanding questions yet to be discussed in this work, as well as some interesting ideas which can be introduced to advance the learning result in the future. We list some of them here:
\begin{itemize}[leftmargin=*]
\item {\em Dependent nonparametric priors:} since our reward function depends on past rewards, a dependent prior which incorporates result learned in previous iteration can propel the convergence of posterior inference by empowering the agent to better exploit the accumulated knowledge.
\item {\em Uneven priorities for agents:} we only considered equal priorities for all agents in this work, however, in real-world application wireless nodes switch between different priority levels. If some agents belong to different priority groups, weighting factor may be necessary in the learning process to reflect the changing priorities for different agents.
\item {\em Joint optimization for global and local interests:} in our model there is only one global reward for the agent team to optimize. If each agent can observe its local reward, the multi-agent reinforcement learning becomes multi-task, where all agent optimize joint policy for global value and each agent optimizes its local policy for local value simultaneously.
\item {\em Different design of reward function:} the design of reward function encodes the core objective of the learning process, with different performance measurement there can be different designs for the reward function.
\item {\em Simulation with large data set:} due to the lack of computing resource, we only simulated our model with small dataset. More complete simulations with larger dataset and scale should be performed to demonstrate the robustness of our algorithm.
\end{itemize}







\bibliography{References}

\begin{thebibliography}{10}

\bibitem{AmatoFSC}
C.~Amato, D.~Bernstein, and S.~Zilberstein.
\newblock Optimizing fixed-size stochastic controllers for pomdps and
  decentralized pomdps.
\newblock {\em Autonomous Agents and Multi-Agent Systems}, 21:293--320, 11
  2010.

\bibitem{AmatoFSCinit}
C.~Amato and S.~Zilberstein.
\newblock Achieving goals in decentralized pomdps.
\newblock In {\em Proceedings of The 8th International Conference on Autonomous
  Agents and Multiagent Systems - Volume 1}, AAMAS '09, page 593–600,
  Richland, SC, 2009. International Foundation for Autonomous Agents and
  Multiagent Systems.

\bibitem{Q_7925694}
M.~A. Aref, S.~K. Jayaweera, and S.~Machuzak.
\newblock Multi-agent reinforcement learning based cognitive anti-jamming.
\newblock In {\em 2017 IEEE Wireless Communications and Networking Conference},
  pages 1--6, March 2017.

\bibitem{Q_sensors}
R.~Bajracharya, R.~Shrestha, and S.~W. Kim.
\newblock Q-learning based fair and efficient coexistence of lte in unlicensed
  band.
\newblock {\em Sensors (Basel, Switzerland)}, 19(13), June 2019.

\bibitem{bg_5Gunlincesed}
S.~Bayhan, G.~G{\"{u}}r, and A.~Zubow.
\newblock "the future is unlicensed: Coexistence in the unlicensed spectrum for
  5g".
\newblock {\em arXiv preprint arXiv: 1801.04964}, 2018.

\bibitem{Q_5986378}
K.~Chowdhury, R.~Doost-Mohammady, W.~Meleis, M.~D. Felice, and L.~Bononi.
\newblock Cooperation and communication in cognitive radio networks based on tv
  spectrum experiments.
\newblock In {\em 2011 IEEE International Symposium on a World of Wireless,
  Mobile and Multimedia Networks}, pages 1--9, 2011.

\bibitem{cisco}
Cisco annual internet report (2018–2023) white paper, March 2020.

\bibitem{3gpp.36.213}
{ETSI TS 36.213}.
\newblock Lte; evolved universal terrestrial radio access (e-utra); physical
  layer procedure.
\newblock Technical Report 36.213, European Telecommunications Standards
  Institute, 2020.
\newblock Version 14.16.0.

\bibitem{ferguson1973}
T.~S. Ferguson.
\newblock A bayesian analysis of some nonparametric problems.
\newblock {\em The annals of statistics}, pages 209--230, 1973.

\bibitem{guo2018explaining}
W.~Guo, S.~Huang, Y.~Tao, X.~Xing, and L.~Lin.
\newblock Explaining deep learning models--a bayesian non-parametric approach.
\newblock In {\em Advances in Neural Information Processing Systems}, pages
  4514--4524, 2018.

\bibitem{9093212}
M.~{Han}, S.~{Khairy}, L.~X. {Cai}, Y.~{Cheng}, and R.~{Zhang}.
\newblock Reinforcement learning for efficient and fair coexistence between
  lte-laa and wi-fi.
\newblock {\em IEEE Transactions on Vehicular Technology}, 69(8):8764--8776,
  2020.

\bibitem{bg_HAN201653}
Y.~Han, E.~Ekici, H.~Kremo, and O.~Altintas.
\newblock Spectrum sharing methods for the coexistence of multiple rf systems:
  A survey.
\newblock {\em Ad Hoc Networks}, 53:53 -- 78, 2016.

\bibitem{HansenFSC}
E.~A. Hansen.
\newblock Solving pomdps by searching in policy space.
\newblock In {\em Proceedings of the Fourteenth Conference on Uncertainty in
  Artificial Intelligence}, UAI'98, page 211–219, San Francisco, CA, USA,
  1998. Morgan Kaufmann Publishers Inc.

\bibitem{jain1984fairness}
R.~K. Jain, D.-M.~W. Chiu, and W.~R. Hawe.
\newblock A quantitative measure of fairness and discrimination.
\newblock {\em Eastern Research Laboratory, Digital Equipment Corporation,
  Hudson, MA}, 1984.

\bibitem{kota}
J.~Kota, G.~Jacyna, and A.~Papandreou-Suppappola.
\newblock Nonstationary signal design for coexisting radar and communications
  systems.
\newblock In {\em 50th Asilomar Conference on Signals, Systems and Computers},
  pages 549--553, 2016.

\bibitem{kumarDBN}
A.~Kumar and S.~Zilberstein.
\newblock Anytime planning for decentralized pomdps using expectation
  maximization.
\newblock {\em arXiv preprint arXiv:1203.3490}, 2012.

\bibitem{6417543}
Z.~Lan, H.~Jiang, and X.~Wu.
\newblock Decentralized cognitive mac protocol design based on pomdp and
  q-learning.
\newblock In {\em 7th International Conference on Communications and Networking
  in China}, pages 548--551, 2012.

\bibitem{6415750}
S.~Lee, S.~Park, G.~Noh, Y.~Park, and D.~Hongt.
\newblock Energy-efficient spectrum access for ultra low power sensor networks.
\newblock In {\em MILCOM 2012 - 2012 IEEE Military Communications Conference},
  pages 1--6, 2012.

\bibitem{li09multi}
H.~Li, X.~Liao, and L.~Carin.
\newblock Multi-task reinforcement learning in partially observable stochastic
  environments.
\newblock {\em Journal of Machine Learning Research}, 10(5), 2009.

\bibitem{9069218}
L.~{Li}, L.~{Liu}, J.~{Bai}, H.~H. {Chang}, H.~{Chen}, J.~D. {Ashdown},
  J.~{Zhang}, and Y.~{Yi}.
\newblock Accelerating model-free reinforcement learning with imperfect model
  knowledge in dynamic spectrum access.
\newblock {\em IEEE Internet of Things Journal}, 7(8):7517--7528, 2020.

\bibitem{liu_thesis}
M.~Liu.
\newblock {\em Efficient Bayesian Nonparametric Methods for Model-Free
  Reinforcement Learning in Centralized and Decentralized Sequential
  Environments}.
\newblock PhD thesis, Duke University, 2016.

\bibitem{LiuSBPR}
M.~Liu, C.~Amato, X.~Liao, L.~Carin, and J.~P. How.
\newblock Stick-breaking policy learning in dec-pomdps.
\newblock In {\em 24th International Joint Conference on Artificial
  Intelligence (IJCAI 2015)}, pages 2011--2018, 2015.

\bibitem{Q_8645080}
O.~Ma, A.~R. Chiriyath, A.~Herschfelt, and D.~W. Bliss.
\newblock Cooperative radar and communications coexistence using reinforcement
  learning.
\newblock In {\em 2018 52nd Asilomar Conference on Signals, Systems, and
  Computers}, pages 947--951, 2018.

\bibitem{Q_fairCoex}
V.~Maglogiannis, D.~Naudts, A.~Shahid, , and I.~Moerman.
\newblock A q-learning scheme for fair coexistence between lte and wi-fi in
  unlicensed spectrum.
\newblock {\em IEEE Access}, 6:27278--27293, 2018.

\bibitem{DBLPBahman}
B.~Moraffah.
\newblock Inference for multiple object tracking: {A} bayesian nonparametric
  approach.
\newblock {\em CoRR}, abs/1909.06984, 2019.

\bibitem{moraffah2019inference}
B.~Moraffah.
\newblock Inference for multiple object tracking: A bayesian nonparametric
  approach.
\newblock {\em arXiv preprint arXiv:1909.06984}, 2019.

\bibitem{moraffah2019tracking}
B.~Moraffah, C.~Brito, B.~Venkatesh, and A.~Papandreou-Suppappola.
\newblock Tracking multiple objects with multimodal dependent measurements:
  Bayesian nonparametric modeling.
\newblock pages 1847--1851, 2019.

\bibitem{moraffah2019use}
B.~Moraffah, C.~Brito, B.~Venkatesh, and A.~Papandreou-Suppappola.
\newblock Use of hierarchical dirichlet processes to integrate dependent
  observations from multiple disparate sensors for tracking.
\newblock In {\em 2019 22th International Conference on Information Fusion
  (FUSION)}, pages 1--7. IEEE, 2019.

\bibitem{moraffah2018dependent}
B.~Moraffah and A.~Papandreou-Suppappola.
\newblock Dependent dirichlet process modeling and identity learning for
  multiple object tracking.
\newblock In {\em 2018 52nd Asilomar Conference on Signals, Systems, and
  Computers}, pages 1762--1766. IEEE, 2018.

\bibitem{moraffah2019random}
B.~Moraffah and A.~Papandreou-Suppappola.
\newblock Random infinite tree and dependent poisson diffusion process for
  nonparametric bayesian modeling in multiple object tracking.
\newblock In {\em ICASSP 2019-2019 IEEE International Conference on Acoustics,
  Speech and Signal Processing (ICASSP)}, pages 5217--5221. IEEE, 2019.

\bibitem{moraffah2019nonparametric}
B.~Moraffah, A.~Papandreou-Suppappola, and M.~Rangaswamy.
\newblock Nonparametric bayesian methods and the dependent pitman-yor process
  for modeling evolution in multiple object tracking.
\newblock In {\em 2019 22th International Conference on Information Fusion
  (FUSION)}, pages 1--6. IEEE, 2019.

\bibitem{moraffah2021clutter}
B.~Moraffah, C.~Richmond, R.~Moraffah, and A.~Papandreou-Suppappola.
\newblock Use of bayesian nonparametric methods for estimating the measurements
  in high clutter.
\newblock In {\em CoRR abs/2012.09785 (2020)}. arXiv, 2020.

\bibitem{POMDPbook}
F.~A. Oliehoek and C.~Amato.
\newblock {\em A Concise Introduction to Decentralized POMDPs}.
\newblock Springer Publishing Company, Incorporated, 1st edition, 2016.

\bibitem{6482133}
J.~Pajarinen, A.~Hottinen, and J.~Peltonen.
\newblock Optimizing spatial and temporal reuse in wireless networks by
  decentralized partially observable markov decision processes.
\newblock {\em IEEE Transactions on Mobile Computing}, 13(4):866--879, 2014.

\bibitem{bg_wifiLTEDC}
Y.~Pang, A.~Babaei, J.~Andreoli-Fang, and B.~Hamzeh.
\newblock Wi-fi coexistence with duty cycled {LTE-U}.
\newblock {\em CoRR}, abs/1606.07972, 2016.

\bibitem{pitman2002GEM}
J.~Pitman.
\newblock Combinatorial stochastic processes.
\newblock Technical report, Technical Report 621, Dept. Statistics, UC
  Berkeley, 2002. Lecture notes for St. Flour Summer School, 2002.

\bibitem{DBLPPolatkan}
G.~Polatkan, M.~Zhou, L.~Carin, D.~M. Blei, and I.~Daubechies.
\newblock A bayesian nonparametric approach to image super-resolution.
\newblock {\em CoRR}, abs/1209.5019, 2012.

\bibitem{RL_8378616}
E.~Selvi, R.~M. Buehrer, A.~Martone, and K.~Sherbondy.
\newblock On the use of markov decision processes in cognitive radar: An
  application to target tracking.
\newblock In {\em 2018 IEEE Radar Conference}, pages 0537--0542, April 2018.

\bibitem{802.11n_cisco}
I.~C. Society.
\newblock Ieee standard for information technology--local and metropolitan area
  networks--specific requirements--part 11: Wireless lan medium access control
  (mac) band physical layer (phy) specifications amendment 5: Enhancements for
  higher throughput.
\newblock {\em IEEE Std 802.11n-2009 (Amendment to IEEE Std 802.11-2007 as
  amended by IEEE Std 802.11k-2008, IEEE Std 802.11r-2008, IEEE Std
  802.11y-2008, and IEEE Std 802.11w-2009)}, pages 1--565, 2009.

\bibitem{exmethod_6503914}
S.~Sodagari, A.~Khawar, T.~C. Clancy, and R.~McGwier.
\newblock A projection based approach for radar and telecommunication systems
  coexistence.
\newblock In {\em 2012 IEEE Global Communications Conference (GLOBECOM)}, pages
  5010--5014, 2012.

\bibitem{8288850}
Y.~{Su}, X.~{Du}, L.~{Huang}, Z.~{Gao}, and M.~{Guizani}.
\newblock Lte-u and wi-fi coexistence algorithm based on q-learning in
  multi-channel.
\newblock {\em IEEE Access}, 6:13644--13652, 2018.

\bibitem{RLbook}
R.~S. Sutton and A.~G. Barto.
\newblock {\em Reinforcement Learning: An Introduction}.
\newblock MIT Press, Cambridge, MA, USA, 1998.

\bibitem{toussaintDBN}
M.~Toussaint and A.~Storkey.
\newblock Probabilistic inference for solving discrete and continuous state
  markov decision processes.
\newblock In {\em Proceedings of the 23rd international conference on Machine
  learning}, pages 945--952, 2006.

\bibitem{8646702}
T.~{Tsiligkaridis} and D.~{Romero}.
\newblock Reinforcement learning with budget-constrained nonparametric function
  approximation for opportunistic spectrum access.
\newblock In {\em 2018 IEEE Global Conference on Signal and Information
  Processing (GlobalSIP)}, pages 579--583, 2018.

\bibitem{7249182}
Y.~Xiao, Z.~Han, D.~Niyato, and C.~Yuen.
\newblock Bayesian reinforcement learning for energy harvesting communication
  systems with uncertainty.
\newblock In {\em 2015 IEEE International Conference on Communications (ICC)},
  pages 5398--5403, 2015.

\bibitem{9064881}
Z.~{Yan}, P.~{Cheng}, Z.~{Chen}, Y.~{Li}, and B.~{Vucetic}.
\newblock Gaussian process reinforcement learning for fast opportunistic
  spectrum access.
\newblock {\em IEEE Transactions on Signal Processing}, 68:2613--2628, 2020.

\bibitem{Q_dynchanselCR}
K.-L.~A. Yau, P.~Komisarczuk, and P.~D. Teal.
\newblock A context-aware and intelligent dynamic channel selection scheme for
  cognitive radio networks.
\newblock In {\em 2009 4th International Conference on Cognitive Radio Oriented
  Wireless Networks and Communications}, pages 1--6, June 2009.

\bibitem{pmlr-zhang18j}
A.~Zhang and J.~Paisley.
\newblock Deep bayesian nonparametric tracking.
\newblock In J.~Dy and A.~Krause, editors, {\em Proceedings of the 35th
  International Conference on Machine Learning}, volume~80 of {\em Proceedings
  of Machine Learning Research}, pages 5833--5841, Stockholmsmässan, Stockholm
  Sweden, 10--15 July 2018. PMLR.

\bibitem{LTEWiFisurvey}
S.~Zinno, G.~Stasi, S.~Avallone, and G.~Ventre.
\newblock On a fair coexistence of lte and wi-fi in the unlicensed spectrum: A
  survey.
\newblock {\em Computer Communications}, 115:35--50, 11 2018.

\end{thebibliography}
\bibliographystyle{abbrv}

\newpage
\appendix

\section{Appendix}

\subsection{Empirical Value Function}
\label{ch: em_value}
To prove that $p(a_{0:t}|o_{1:t})=\prod_{\tau=0}^{t}p(a_{\tau}|h_{\tau})=\prod_{\tau=0}^{t}p(a_{\tau}|a_{0:\tau},o_{1:\tau-1})$, we expand
\begin{equation}
\begin{aligned}
    &p(a_{0:t}|o_{1:t}) \\
    &=\sum_{z_0=1}^{|Z|}\cdots\sum_{z_t=1}^{|Z|}p(a_{0:t},z_{0:t}|o_{1:t})\\
    &=\sum_{z_0=1}^{|Z|}\cdots\sum_{z_t=1}^{|Z|}p(z_0)p(a_0|z_0)\prod_{\tau=1}^{t}p(z_{\tau}|z_{\tau-1},a_{\tau-1},o_{\tau})p(a_{\tau}|z_{\tau})
\end{aligned}
\end{equation}
And since observation $o_t$ does not influence action before time $t$,
\begin{equation}
\begin{aligned}
    &p(a_{0:t-1}|o_{1:t}) \\
    &=\sum_{z_0=1}^{|Z|}\cdots\sum_{z_t=1}^{|Z|}\sum_{a_t=1}^{|A|}p(a_t,a_{0:t-1},z_{0:t}|o_{1:t})\\
    &=\sum_{z_0,...,z_{t}=1}^{|Z|}\sum_{a_t=1}^{|A|}\left[p(z_0)p(a_0|z_0)\prod_{\tau=1}^{t-1}p(z_{\tau}|z_{\tau-1},a_{\tau-1},o_{\tau})p(a_{\tau}|z_{\tau})\right]\\
    &\times p(z_{t}|z_{t-1},a_{t-1},o_{t})p(a_t|z_t)\\
    &=\sum_{z_0,...,z_{t-1}=1}^{|Z|}\left[p(z_0)p(a_0|z_0)\prod_{\tau=1}^{t-1}p(z_{\tau}|z_{\tau-1},a_{\tau-1},o_{\tau})p(a_{\tau}|z_{\tau})\right] \\
    &\times\sum_{z_t=1}^{|Z|}\sum_{a_t=1}^{|A|}p(z_{t}|z_{t-1},a_{t-1},o_{t})p(a_t|z_t)\\
    &=\sum_{z_0,...,z_{t-1}=1}^{|Z|}\left[p(z_0)p(a_0|z_0)\prod_{\tau=1}^{t-1}p(z_{\tau}|z_{\tau-1},a_{\tau-1},o_{\tau})p(a_{\tau}|z_{\tau})\right]\\
    &=\sum_{z_0,...,z_{t-1}=1}^{|Z|}p(a_{0:t-1},z_{0:t-1}|o_{1:t-1})\\
    &=p(a_{0:t-1}|o_{1:t-1})
\end{aligned}
\label{equ: B2}
\end{equation}
Decompose each $p(a_{\tau}|h_{\tau})$ as
\begin{equation}
    p(a_{\tau}|h_{\tau})=p(a_{\tau}|a_{0:\tau-1},o_{1:\tau})=\frac{p(a_{0:\tau}|o_{1:\tau})}{p(a_{0:\tau-1}|o_{1:\tau})}=\frac{p(a_{0:\tau}|o_{1:\tau})}{p(a_{0:\tau-1}|o_{1:\tau-1})}
\label{equ: B3}
\end{equation}
Combine \ref{equ: B2} and \ref{equ: B3}, there is
\begin{equation}
\begin{aligned}
    &\prod_{\tau=0}^{t}p(a_{\tau}|h_{\tau}) \\
    &=p(a_{t}|a_{0:t-1},o_{1:t})p(a_{t-1}|a_{0:t-2},o_{1:t-1})\cdots p(a_1|a_0,o_1)p(a_0)\\
    &=\frac{p(a_{0:t}|o_{1:t})}{p(a_{0:t-1}|o_{1:t-1})}\frac{p(a_{0:t-1}|o_{1:t-1})}{p(a_{0:t-2}|o_{1:t-2})}\cdots\frac{p(a_{0:1}|o_{1})}{p(a_{0})}p(a_0)\\
    &=p(a_{0:t}|o_{1:t})
\end{aligned}
\end{equation}
\subsection{Computation of Variational Distributions}
\label{ch: VI_update}
Here we provide the proof of \ref{thm: q_update}. The optimal $q$ distribution for each variable can be obtained by taking derivative on $\text{ELBO}(q)$ with respect to the desired $q$ distribution. The $\text{ELBO}(q)$ for our problem has been derived in \ref{equ: ELBO_p} to \ref{equ: ELBO_final}. By taking derivative on \ref{equ: ELBO_final} with respect to each $q\left(z_{n,0:t}^k\right)$, $q(\Theta_n)$, $q(\rho_{n})$, and $q\left(\alpha_{n,a,o}^{i}\right)$ respectively while keeping all others fixed then reorganize it in terms of the distribution form defined in \ref{equ: q_dist}, each optimal $q$ distribution can be computed analytically.

For $q\left(z_{n,0:t}^k\right)$, keep all $q(\Theta_{n})$, $q(\rho_n)$, and $q\left(\alpha_{n,a,o}^{i}\right)$ fixed, the optimal $q^*\left(z_{n,0:t}^k\right)$ is obtained from $\frac{\partial}{\partial q\left(z_{n,t}^k\right)}\text{ELBO}(q)=0$ with constraint
\begin{equation}
    \sum_{k=1}^K\frac{1}{K}\sum_{t=0}^{T_k}\sum_{z_{1:N,0}^k=1}^{|\mathcal{Z}|}\cdots\sum_{z_{1:N,t}^k=1}^{|\mathcal{Z}|}\prod_{n=1}^{N}q\left(z_{n,0:t}^{k}\right)=1, \, \forall (n,k,t) \text{ indices}
\label{equ:q(z)_constraint}
\end{equation}
we have
\begin{equation}
\begin{aligned}
&\frac{\partial}{\partial q\left(z_{n,t}^k\right)}\text{ELBO}(q) \\
&=\sum_{k=1}^{K}\frac{1}{K}\sum_{t=0}^{T_k}\sum_{\Vec{z}_{0}^{k}...\Vec{z}_{t}^{k}=1}^{|\mathcal{Z}|}\int\prod_{i\neq n}q\left(z_{i,0:t}^{k}\right)q(\Theta)\ln\Tilde{r}_{t}^{k}\prod_{n=1}^{N}p\left(a_{n,0:t}^{k},z_{n,0:t}^{k}\vert o_{n,1:t}^{k},\Theta\right)d\Theta \\
&-\sum_{k=1}^{K}\frac{1}{K}\sum_{t=0}^{T_k}\sum_{\Vec{z}_{0}^{k}...\Vec{z}_{t}^{k}=1}^{|\mathcal{Z}|}\prod_{i\neq n}q\left(z_{i,0:t}^{k}\right)\left[\ln\prod_{i=1}^{N}q\left(z_{i,0:t}^{k}\right)\right] \\
&-\sum_{k=1}^{K}\frac{1}{K}\sum_{t=0}^{T_k}\sum_{\Vec{z}_{0}^{k}...\Vec{z}_{t}^{k}=1}^{|\mathcal{Z}|}\prod_{i\neq n}q\left(z_{i,0:t}^{k}\right) \\
&=0
\end{aligned}
\end{equation}
$q(\rho)$ and $q(\alpha)$ integrate out in above equation since they are not directly associated to $z_{n,t}^k$. Remove all terms unrelated to $q(z_{n,t}^k)$ and rearrange terms, we obtain
\begin{equation}
\begin{aligned}
&\sum_{k=1}^{K}\frac{1}{K}\sum_{t=0}^{T_k}\sum_{\Vec{z}_{0}^{k}...\Vec{z}_{t}^{k}=1}^{|\mathcal{Z}|}\int\prod_{i\neq n}q(z_{i,0:t}^{k})q(\Theta)\ln\Tilde{r}_{t}^{k}\prod_{n=1}^{N}p\left(a_{n,0:t}^{k},z_{n,0:t}^{k}\vert o_{n,1:t}^{k},\Theta\right)d\Theta \\
&=\sum_{k=1}^{K}\frac{1}{K}\sum_{t=0}^{T_k}\sum_{\Vec{z}_{0}^{k}...\Vec{z}_{t}^{k}=1}^{|\mathcal{Z}|}\prod_{i\neq n}q\left(z_{i,0:t}^{k}\right)\left[\ln q\left(z_{n,0:t}^{k}\right)\right] \\
&\rightarrow \sum_{k=1}^{K}\frac{1}{K}\sum_{t=0}^{T_k}\sum_{\Vec{z}_{0}^{k}...\Vec{z}_{t}^{k}=1}^{|\mathcal{Z}|}\prod_{i\neq n}q\left(z_{i,0:t}^{k}\right)\left[\int q(\Theta)\ln\Tilde{r}_{t}^{k}\prod_{n=1}^{N}p
\left(a_{n,0:t}^{k},z_{n,0:t}^{k}\vert o_{n,1:t}^{k},\Theta\right)d\Theta\right] \\
&=\sum_{k=1}^{K}\frac{1}{K}\sum_{t=0}^{T_k}\sum_{\Vec{z}_{0}^{k}...\Vec{z}_{t}^{k}=1}^{|\mathcal{Z}|}\prod_{i\neq n}q\left(z_{i,0:t}^{k}\right)\left[\ln q
\left(z_{n,0:t}^{k}\right)\right] \\
&\rightarrow \ln q\left(z_{n,0:t}^{k}\right) \\
&=\int q(\Theta)\ln\Tilde{r}_{t}^{k}\prod_{n=1}^{N}p\left(a_{n,0:t}^{k},z_{n,0:t}^{k}\vert o_{n,1:t}^{k},\Theta\right)d\Theta d\alpha \\
&=\text{E}_{q(\Theta)}\left[\ln\Tilde{r}_{t}^{k}\prod_{n=1}^{N}p\left(a_{n,0:t}^{k},z_{n,0:t}^{k}\vert o_{n,1:t}^{k},\Theta\right)\right]
\end{aligned}
\end{equation}
The optimal $q^*\left(z_{n,0:t}^{k}\right)$ has the form
\begin{equation}
\begin{aligned}
& q^*\left(z_{n,0:t}^{k}\right) \\
&\propto \exp\left\{\text{E}_{q(\Theta)}\left[\ln\Tilde{r}_{t}^{k}\prod_{n=1}^{N}p\left(a_{n,0:t}^{k},z_{n,0:t}^{k}\vert o_{n,1:t}^{k},\Theta\right)\right]\right\} \\
&=\exp\left\{\text{E}_{q(\Theta)}\left[\ln\Tilde{r}_{t}^{k}\right]+\sum_{n=1}^{N}\text{E}_{q(\Theta)}\left[\ln p\left(a_{n,0:t}^{k},z_{n,0:t}^{k}\vert o_{n,1:t}^{k},\Theta\right)\right]\right\}
\end{aligned}
\label{equ:q(z)}
\end{equation}
Due to the independence between agents, remove all term with indices other than $(n,k,t)$, the above equation is proportional to
\begin{equation}
\medmuskip=1mu
\thinmuskip=1mu
\thickmuskip=1mu
\begin{aligned}
&\exp\left\{\text{E}_{q(\Theta)}\left[\ln\Tilde{r}_{t}^{k}\right]+\text{E}_{q(\Theta)}\left[\ln p\left(a_{n,0:t}^{k},z_{n,0:t}^{k}\vert o_{n,1:t}^{k},\Theta\right)\right]\right\} \\
&=\exp\left\{\ln\Tilde{r}_{t}^{k}+\text{E}_{q(\Theta)}\left[\ln p\left(a_{n,0:t}^{k},z_{n,0:t}^{k}\vert o_{n,1:t}^{k},\Theta\right)\right]\right\} \\
&=\Tilde{r}_{t}^{k}\exp\left\{\text{E}_{q(\Theta)}\left[\ln \eta_n^{z_0}\pi_{n,z_0}^{k,a_0}\prod_{\tau=1}^{t}\omega_{n,a_{\tau-1},o_{\tau}}^{k,z_{\tau-1},z_{\tau}}\pi_{n,z_{\tau}}^{k,a_{\tau}}\right]\right\} \\
&=\Tilde{r}_{t}^{k}\exp\left\{\text{E}_{q(\Theta)}\left[\ln \eta_n^{z_0}+\sum_{\tau=0}^{t}\ln\pi_{n,z_{\tau}}^{k,a_{\tau}}+\sum_{\tau=1}^{t}\ln\omega_{n,a_{\tau-1},o_{\tau}}^{k,z_{\tau-1},z_{\tau}}\right]\right\} \\
&=\Tilde{r}_{t}^{k}\exp\left\{\text{E}_{q(u)}\left[\ln\eta_n^{z_0}\right]+\sum_{\tau=0}^{t}\text{E}_{q(\pi)}\left[\ln\pi_{n,z_{\tau}}^{k,a_{\tau}}\right]+\sum_{\tau=1}^{t}\text{E}_{q(V)}\left[\ln\omega_{n,a_{\tau-1},o_{\tau}}^{k,z_{\tau-1},z_{\tau}}\right]\right\} \\
&=\Tilde{r}_{t}^{k}\exp\left\{\text{E}_{q(u)}\left[\ln\eta_n^{z_0}\right]\right\}\prod_{\tau=0}^{t}\exp\left\{\text{E}_{q(\pi)}\left[\ln\pi_{n,z_{\tau}}^{k,a_{\tau}}\right]\right\}\prod_{\tau=1}^{t}\exp\left\{\text{E}_{q(V)}\left[\ln\omega_{n,a_{\tau-1},o_{\tau}}^{k,z_{\tau-1},z_{\tau}}\right]\right\} \\
&=\Tilde{r}_{t}^{k}\Tilde{\eta}_n^{z_0}\prod_{\tau=0}^{t}\Tilde{\pi}_{n,z_{n,\tau}^{k}}^{a_{n,\tau}^{k}}\prod_{\tau=1}^{t}\Tilde{\omega}_{n,a_{\tau-1},o_{\tau}}^{k,z_{\tau-1},z_{\tau}}
\end{aligned}
\label{equ:q(z)_result}
\end{equation}
Where $\Tilde{\Theta}_n=(\Tilde{\eta}_n, \Tilde{\omega}_{n}, \Tilde{\pi}_{n})$ and
\begin{equation}
\begin{aligned}
&\Tilde{\eta}_n^{z_0}=\exp\left\{\text{E}_{q(u)}\left[\ln\eta_n^{z_0}\right]\right\} \\
&\Tilde{\pi}_{n,z_{n,\tau}^{k}}^{a_{n,\tau}^{k}}=\exp\left\{\text{E}_{q(\pi)}\left[\ln\pi_{n,z_{\tau}}^{k,a_{\tau}}\right]\right\} =\exp\left\{\Psi\left(\phi_{n,z_{n,\tau}^{k}}^{a_{n,\tau}^{k}}\right)-\Psi\left(\sum_{a=1}^{|\mathcal{A}_n|}\phi_{n,z_{n,\tau}^{k}}^{a}\right)\right\} \\
&\Tilde{\omega}_{n,a_{\tau-1},o_{\tau}}^{k,z_{\tau-1},z_{\tau}}=\exp\left\{\text{E}_{q(V)}\left[\ln\omega_{n,a_{\tau-1},o_{\tau}}^{k,z_{\tau-1},z_{\tau}}\right]\right\}
\end{aligned}
\end{equation}
$\eta$ and $\omega$ are constructed by the stick-breaking process. For different destination node, the terms in exponential $\exp\{\cdot\}$ are computed by 
\begin{align}
&\text{E}_{q(u)}\left[\ln\eta_n^1\right]=\text{E}_{q(u)}\left[\ln u_n^1\right]=\Psi(\delta_n^1)-\Psi(\delta_n^1+\mu_n^1) \\
&\begin{aligned}
&\text{E}_{q(u)}\left[\ln\eta_n^i\right] \\
&=\text{E}_{q(u)}\left[\ln u_n^i\prod_{m=1}^{i-1}(1-u_n^m)\right] \\
&=\text{E}_{q(u)}\left[\ln u_n^i\right]+\sum_{m=1}^{i-1}\text{E}_{q(u)}\left[\ln (1-u_n^m)\right] \\
&=\Psi(\delta_n^i)-\Psi(\delta_n^i+\mu_n^i)+\sum_{m=1}^{i-1}\left[\Psi(\mu_n^m)-\Psi(\delta_n^m+\mu_n^m)\right] \ \text{for }i=2,...,|\mathcal{Z}_n|-1
\end{aligned} \\
&\text{E}_{q(u)}\left[\ln\eta_n^{|\mathcal{Z}_n|}\right]=\text{E}_{q(u)}\left[\ln \prod_{m=1}^{|\mathcal{Z}_n|-1}(1-u_n^m)\right]=\sum_{m=1}^{|\mathcal{Z}_n|-1}\left[\Psi(\mu_n^m)-\Psi(\delta_n^m+\mu_n^m)\right]
\end{align}
and
\begin{align}
&\begin{aligned}
&\text{E}_{q(V)}\left[\ln\omega_{n,a_{\tau-1},o_{\tau}}^{k,z_{\tau-1},1}\right] \\
&=\text{E}_{q(V)}\left[\ln V_{n,a_{\tau-1},o_{\tau}}^{k,z_{\tau-1},1}\right] \\ &=\Psi\left(\sigma_{n,a_{\tau-1},o_{\tau}}^{k,z_{\tau-1},1}\right)-\Psi\left(\sigma_{n,a_{\tau-1},o_{\tau}}^{k,z_{\tau-1},1}+\lambda_{n,a_{\tau-1},o_{\tau}}^{k,z_{\tau-1},1}\right)
\end{aligned} \\
&\begin{aligned}
&\text{E}_{q(V)}\left[\ln\omega_{n,a_{\tau-1},o_{\tau}}^{k,z_{\tau-1},i}\right] \\
&=\text{E}_{q(V)}\left[\ln V_{n,a_{\tau-1},o_{\tau}}^{k,z_{\tau-1},i}\prod_{m=1}^{i-1}\left(1-V_{n,a_{\tau-1},o_{\tau}}^{k,z_{\tau-1},m}\right)\right] \\
&=\Psi\left(\sigma_{n,a_{\tau-1},o_{\tau}}^{k,z_{\tau-1},i}\right)-\Psi\left(\sigma_{n,a_{\tau-1},o_{\tau}}^{k,z_{\tau-1},i}+\lambda_{n,a_{\tau-1},o_{\tau}}^{k,z_{\tau-1},i}\right) \\
&+\sum_{m=1}^{i-1}\left[\Psi\left(\lambda_{n,a_{\tau-1},o_{\tau}}^{k,z_{\tau-1},m}\right)-\Psi\left(\sigma_{n,a_{\tau-1},o_{\tau}}^{k,z_{\tau-1},m}+\lambda_{n,a_{\tau-1},o_{\tau}}^{k,z_{\tau-1},m}\right)\right] \ \text{for }i=2,...,|\mathcal{Z}_n|-1
\end{aligned} \\
&\begin{aligned}
&\text{E}_{q(V)}\left[\ln\omega_{n,a_{\tau-1},o_{\tau}}^{k,z_{\tau-1},{|\mathcal{Z}_n|}}\right] \\
&=\text{E}_{q(V)}\left[\ln \prod_{m=1}^{|\mathcal{Z}_n|-1}\left(1-V_{n,a_{\tau-1},o_{\tau}}^{k,z_{\tau-1},m}\right)\right] \\
&=\sum_{m=1}^{|\mathcal{Z}_n|-1}\left[\Psi\left(\lambda_{n,a_{\tau-1},o_{\tau}}^{k,z_{\tau-1},m}\right)-\Psi\left(\sigma_{n,a_{\tau-1},o_{\tau}}^{k,z_{\tau-1},m}+\lambda_{n,a_{\tau-1},o_{\tau}}^{k,z_{\tau-1},m}\right)\right]
\end{aligned}
\end{align}
In \ref{equ:q(z)}, the proportional expression is utilized to represent the $q\left(z_{n,0:t}^{k}\right)$. To make $q\left(z_{n,0:t}^{k}\right)$ proper distribution of $z_{n,0:t}^{k}$, i.e., satisfy \ref{equ:q(z)_constraint}, we re-write the final result in \ref{equ:q(z)_result} and substitute it into the constraint equation
\begin{equation}
\medmuskip=1mu
\thinmuskip=1mu
\thickmuskip=1mu
\begin{aligned}
&\frac{1}{K}\sum_{k,t}\sum_{z_{1:N,0}^k=1}^{|\mathcal{Z}|}\cdots\sum_{z_{1:N,t}^k=1}^{|\mathcal{Z}|}\prod_{n=1}^{N}q\left(z_{n,0:t}^{k}\right) \\
&=\frac{1}{K}\sum_{k,t}\sum_{z_{1:N,0}^k=1}^{|\mathcal{Z}|}\cdots\sum_{z_{1:N,t}^k=1}^{|\mathcal{Z}|}\prod_{n=1}^{N}\Tilde{r}_{t}^{k}\Tilde{\eta}_n^{z_0}\prod_{\tau=0}^{t}\Tilde{\pi}_{n,z_{n,\tau}^{k}}^{a_{n,\tau}^{k}}\prod_{\tau=1}^{t}\Tilde{\omega}_{n,a_{\tau-1},o_{\tau}}^{k,z_{\tau-1},z_{\tau}} \\
&=\frac{1}{K}\sum_{k,t}\Tilde{r}_{t}^{k}\sum_{z_{1:N,0}^k=1}^{|\mathcal{Z}|}\cdots\sum_{z_{1:N,t}^k=1}^{|\mathcal{Z}|}\prod_{n=1}^{N}p\left(a_{n,0:t}^k,z_{n,0:t}^k\vert o_{n,1:t}^k,\Tilde{\Theta}_n\right) \\
&=\frac{1}{K}\sum_{k,t}\Tilde{r}_{t}^{k}\sum_{z_{1:N,0}^k=1}^{|\mathcal{Z}|}\cdots\sum_{z_{1:N,t}^k=1}^{|\mathcal{Z}|}\prod_{n=1}^{N}p\left(a_{n,0:t}^k\vert o_{n,1:t}^k,\Tilde{\Theta}_n\right)p\left(z_{n,0:t}^k\vert a_{n,0:t}^k,o_{n,1:t}^k,\Tilde{\Theta}_n\right) \\
&=\frac{1}{K}\sum_{k,t}\Tilde{r}_{t}^{k}\sum_{z_{1:N,0}^k=1}^{|\mathcal{Z}|}\cdots\sum_{z_{1:N,t}^k=1}^{|\mathcal{Z}|}\prod_{n=1}^{N}p\left(a_{n,0:t}^k\vert o_{n,1:t}^k,\Tilde{\Theta}_n\right)\prod_{n=1}^{N}p\left(z_{n,0:t}^k\vert a_{n,0:t}^k,o_{n,1:t}^k,\Tilde{\Theta}_n\right) \\
&=\frac{1}{K}\sum_{k,t}\Tilde{r}_{t}^{k}\prod_{n=1}^{N}p\left(a_{n,0:t}^k\vert o_{n,1:t}^k,\Tilde{\Theta}_n\right)\sum_{z_{1:N,0}^k=1}^{|\mathcal{Z}|}\cdots\sum_{z_{1:N,t}^k=1}^{|\mathcal{Z}|}\prod_{n=1}^{N}p\left(z_{n,0:t}^k\vert a_{n,0:t}^k,o_{n,1:t}^k,\Tilde{\Theta}_n\right) \\
&=\frac{1}{K}\sum_{k,t}\Tilde{r}_{t}^{k}\prod_{n=1}^{N}p\left(a_{n,0:t}^k\vert o_{n,1:t}^k,\Tilde{\Theta}_n\right)
\end{aligned}
\end{equation}
By \ref{def: emp_likelihood} and $\Tilde{r}_t^k=\gamma^t\frac{r_t^k-R_{\text{min}}}{\prod_{n=1}^{N}p(a_{n,0:t}^k|o_{n,1:t}^k,\Pi)}$, the above result is just equal to $\hat{V}(D^K;\Tilde{\Theta})$. Thus,
\begin{equation}
\medmuskip=1mu
\thinmuskip=1mu
\thickmuskip=1mu
\begin{aligned}
&\frac{1}{K}\sum_{k,t}\frac{\Tilde{r}_{t}^{k}\prod_{n=1}^{N}p\left(a_{n,0:t}^k\vert o_{n,1:t}^k,\Tilde{\Theta}_n\right)}{\hat{V}(D^K;\Tilde{\Theta})}\sum_{z_{1:N,0}^k=1}^{|\mathcal{Z}|}\cdots\sum_{z_{1:N,t}^k=1}^{|\mathcal{Z}|}\prod_{n=1}^{N}p\left(z_{n,0:t}^k\vert a_{n,0:t}^k,o_{n,1:t}^k,\Tilde{\Theta}_n\right) \\
&=\frac{1}{K}\sum_{k,t}\Tilde{\nu}_{t}^{k}\sum_{z_{1:N,0}^k=1}^{|\mathcal{Z}|}\cdots\sum_{z_{1:N,t}^k=1}^{|\mathcal{Z}|}\prod_{n=1}^{N}p\left(z_{n,0:t}^k\vert a_{n,0:t}^k,o_{n,1:t}^k,\Tilde{\Theta}_n\right) \\
&=\frac{1}{K}\sum_{k,t}\sum_{z_{1:N,0}^k=1}^{|\mathcal{Z}|}\cdots\sum_{z_{1:N,t}^k=1}^{|\mathcal{Z}|}\prod_{n=1}^{N}\Tilde{\nu}_{t}^{k}p\left(z_{n,0:t}^k\vert a_{n,0:t}^k,o_{n,1:t}^k,\Tilde{\Theta}_n\right) \\
&=\frac{1}{K}\sum_{k,t}\sum_{z_{1:N,0}^k=1}^{|\mathcal{Z}|}\cdots\sum_{z_{1:N,t}^k=1}^{|\mathcal{Z}|}\prod_{n=1}^{N}q\left(z_{n,0:t}^k\right) \\
&=1
\end{aligned}
\end{equation}
$\Tilde{\nu}_{t}^{k}$, defined in \ref{equ: q_dist}, is the reweighted reward that makes $q\left(z_{n,0:t}^k\right)$ satisfy \ref{equ:q(z)_constraint}.

For optimal $q^*(\Theta_n)$, by taking partial derivative on $\text{ELBO}(q)$ with respect to $q(\Theta_n)$ and treat all terms unrelated to $q(\Theta_n)$ as constants, the result can be obtained as
\begin{equation}
\medmuskip=0mu
\thinmuskip=0mu
\thickmuskip=0mu
\begin{aligned}
&q^*(\Theta_n) \\
&\propto \exp\left\{\text{E}_{q(z,\rho,\alpha)}\left[\ln\Tilde{r}_{t}^{k}p(a_{n,0:t}^{k},z_{n,0:t}^{k}\vert o_{n,1:t}^{k},\Theta)p(\Theta_n)p(\rho_n)p(\alpha_n)\right]\right\} \\
&\propto \exp\left\{\text{E}_{q(z,\rho,\alpha)}\left[\ln\Tilde{r}_{t}^{k}p(a_{n,0:t}^{k},z_{n,0:t}^{k}\vert o_{n,1:t}^{k},\Theta)p(\Theta_n)\right]\right\} \\
&=\exp\left\{\text{E}_{q(z)}\left[\ln\Tilde{r}_{t}^{k}p(a_{n,0:t}^{k},z_{n,0:t}^{k}\vert o_{n,1:t}^{k},\Theta)\right]+\text{E}_{q(\rho,\alpha)}\left[\ln p(\Theta_n)\right]\right\} \\
&=\exp\left\{\text{E}_{q(z)}\left[\ln\Tilde{r}_{t}^{k}\eta_n^{z_0}\pi_{n,z_0}^{k,a_0}\prod_{\tau=1}^{t}\omega_{n,a_{\tau-1},o_{\tau}}^{k,z_{\tau-1},z_{\tau}}\pi_{n,z_{\tau}}^{k,a_{\tau}}\right]+\text{E}_{q(\rho,\alpha)}\left[\ln p(u_n\vert\rho_n)p(V_n\vert\alpha_n)p(\pi_n)\right]\right\} \\
&\propto \exp\Biggl\{\frac{1}{K}\sum_{k,t,z_{n,0:t}^{k}}q(z_{n,0:t}^{k})\left[\ln\eta_n^{z_0}+\sum_{\tau=0}^{t}\ln\pi_{n,z_{\tau}}^{k,a_{\tau}}+\sum_{\tau=1}^{t}\ln\omega_{n,a_{\tau-1},o_{\tau}}^{k,z_{\tau-1},z_{\tau}}\right] \\
&+\text{E}_{q(\rho)}\left[\ln p(u_n\vert\rho_n)\right]+\text{E}_{q(\alpha)}\left[\ln p(V_n\vert\alpha_n)\right]+\ln p(\pi_n)\Biggr\} \\
&=\exp\left\{\left[\frac{1}{K}\sum_{k,t,z_{n,0:t}^{k}}q(z_{n,0:t}^{k})\left[\ln\eta_n^{z_0}\right]+\text{E}_{q(\rho)}\left[\ln p(u_n\vert\rho_n)\right]\right]\right. \\
&+\left[\frac{1}{K}\sum_{k,t,z_{n,0:t}^{k}}q(z_{n,0:t}^{k})\left[\sum_{\tau=1}^{t}\ln\omega_{n,a_{\tau-1},o_{\tau}}^{k,z_{\tau-1},z_{\tau}}\right]+\text{E}_{q(\alpha)}\left[\ln p(V_n\vert\alpha_n)\right]\right] \\
&+\left.\left[\frac{1}{K}\sum_{k,t,z_{n,0:t}^{k}}q(z_{n,0:t}^{k})\left[\sum_{\tau=0}^{t}\ln\pi_{n,z_{\tau}}^{k,a_{\tau}}\right]+\ln p(\pi_n)\right]\right\} \\
&=\exp\left\{\left[\frac{1}{K}\sum_{k,t,z_{n,0:t}^{k}}q(z_{n,0:t}^{k})\left[\ln u_n^{z_0}\prod_{m=1}^{z_0-1}(1-u_n^m)\right]+\text{E}_{q(\rho)}\left[\ln p(u_n\vert\rho_n)\right]\right]\right. \\
&+\left[\frac{1}{K}\sum_{k,t,z_{n,0:t}^{k}}q(z_{n,0:t}^{k})\sum_{\tau=1}^{t}\left[\ln V_{n,a_{\tau-1},o_{\tau}}^{k,z_{\tau-1},z_{\tau}}\prod_{m=1}^{z_{\tau}-1}\left(1-V_{n,a_{\tau-1},o_{\tau}}^{k,z_{\tau-1},m}\right)\right]+\text{E}_{q(\alpha)}\left[\ln p(V_n\vert\alpha_n)\right]\right] \\
&+\left.\left[\frac{1}{K}\sum_{k,t,z_{n,0:t}^{k}}q(z_{n,0:t}^{k})\left[\sum_{\tau=0}^{t}\ln\pi_{n,z_{\tau}}^{k,a_{\tau}}\right]+\ln p(\pi_n)\right]\right\}
\end{aligned}
\end{equation}
In above formula, there are three parts of variables in the exponential term,
{
\medmuskip=0mu
\thinmuskip=0mu
\thickmuskip=0mu
\begin{align}
&\begin{aligned}
&\frac{1}{K}\sum_{k,t,z_{n,0:t}^{k}}q(z_{n,0:t}^{k})\left[\ln u_n^{z_0}\prod_{m=1}^{z_0-1}(1-u_n^m)\right]+\text{E}_{q(\rho)}\left[\ln p(u_n\vert\rho_n)\right] \\
&=\frac{1}{K}\sum_{k,t,z_{n,0:t}^{k}}q(z_{n,0:t}^{k})\left[\ln u_n^{z_0}+\sum_{m=1}^{z_0-1}\ln(1-u_n^m)\right]+\text{E}_{q(\rho)}\left[\ln p(u_n\vert\rho_n)\right]
\end{aligned} \label{equ:eta_update} \\
&\begin{aligned}
&\frac{1}{K}\sum_{k,t,z_{n,0:t}^{k}}q(z_{n,0:t}^{k})\sum_{\tau=1}^{t}\left[\ln V_{n,a_{\tau-1},o_{\tau}}^{k,z_{\tau-1},z_{\tau}}\prod_{m=1}^{z_{\tau}-1}V_{n,a_{\tau-1},o_{\tau}}^{k,z_{\tau-1},m}\right]+\text{E}_{q(\alpha)}\left[\ln p(V_n\vert\alpha_n)\right] \\
&=\frac{1}{K}\sum_{k,t,z_{n,0:t}^{k}}q(z_{n,0:t}^{k})\sum_{\tau=1}^{t}\left[\ln V_{n,a_{\tau-1},o_{\tau}}^{k,z_{\tau-1},z_{\tau}}+\sum_{m=1}^{z_{\tau}-1}\ln\left(1- V_{n,a_{\tau-1},o_{\tau}}^{k,z_{\tau-1},m}\right)\right]+\text{E}_{q(\alpha)}\left[\ln p(V_n\vert\alpha_n)\right]
\end{aligned} \label{equ:omega_update} \\
& \frac{1}{K}\sum_{k,t,z_{n,0:t}^{k}}q(z_{n,0:t}^{k})\left[\sum_{\tau=0}^{t}\ln\pi_{n,z_{\tau}}^{k,a_{\tau}}\right]+\ln p(\pi_n) \label{equ:pi_update}
\end{align}
}
By the conjugacy between prior and likelihood models, we know each $q$ distribution belongs to the same family of its corresponding prior and they are all in exponential family, thus the computation of $q$ distribution can reduce to the computation of its parameters in their exponential expression. By re-positioning components in above equations in terms of each variable, the parameters for each $q$ distribution can be computed.

For $u_n^i$, re-write the prior and variational distributions in terms of exponential family,
\begin{equation}
\begin{aligned}
& \text{E}_{q_(\rho)}\left[\ln p(u_n^i\vert\rho_n)\right]\propto (1-1)\ln u_n^i+\left(\text{E}_{q_(\rho)}\left[\rho_n\right]-1\right)\ln(1-u_n^i) \\
& \ln q(u_n^i)\propto (\delta_n^i-1)\ln u_n^i+(\mu_n^i-1)\ln(1-u_n^i)
\end{aligned}
\end{equation}
For $\ln u_n^i$, only $(z_{n,0}^k=i)$ is associated with it and all cases with indices $m>i$ must be collected for $\ln(1-\ln u_n^i)$; we rearrange components in \ref{equ:eta_update} and obtain
\begin{align}
&\begin{aligned}
&\left[\frac{1}{K}\sum_{k,t,z_{n,0:t}^{k}}q_{n,t}^k(z_{n,0}^{k}=i)\right]\ln u_n^i=(\delta_n^i-1)\ln u_n^i \\
&\rightarrow \delta_n^i=1+\frac{1}{K}\sum_{k,t,z_{n,0:t}^{k}}q_{n,t}^k(z_{n,0}^{k}=i)
\end{aligned} \\
&{
\medmuskip=1mu
\thinmuskip=1mu
\thickmuskip=1mu
\begin{aligned}
&\left[\sum_{m=i+1}^{|\mathcal{Z}_n|-1}\frac{1}{K}\sum_{k,t,z_{n,0:t}^{k}}q_{n,t}^k(z_{n,0}^{k}=m)+\text{E}_{q_(\rho)}\left[\rho_n\right]-1\right]\ln(1-u_n^i)=(\mu_n^i-1)\ln(1-u_n^i) \\
&\rightarrow \mu_n^i=\frac{g_n}{h_n}+\sum_{m=i+1}^{|\mathcal{Z}_n|}\frac{1}{K}\sum_{k,t,z_{n,0:t}^{k}}q_{n,t}^k(z_{n,0}^{k}=i)
\end{aligned}
}
\end{align}
Where $\text{E}_{q_(\rho)}\left[\rho_n\right]=\frac{g_n}{h_n}$.

The update of $q\left(V_{n,a,o}^{i,j}\right)$ is similar to the update of $q(u_n^i)$. Rewrite $q(V_{n,a,o}^{i,j})$ and $\text{E}_{q(\alpha)}\left[p\left(V_{n,a,o}^{i,j}\vert\alpha_{n,a,o}^i\right)\right]$ in terms of exponential family,
\begin{equation}
\medmuskip=1mu
\thinmuskip=1mu
\thickmuskip=1mu
\begin{aligned}
&\text{E}_{q(\alpha)}\left[\ln p\left(V_{n,a,o}^{i,j}\vert\alpha_{n,a,o}^{i}\right)\right]\propto(1-1)\ln V_{n,a,o}^{i,j}+\left(\text{E}_{q(\alpha)}\left[\alpha_{n,a,o}^{i}\right]-1\right)\ln\left(1-V_{n,a,o}^{i,j}\right) \\
&\ln q\left(V_{n,a,o}^{i,j}\right)\propto \left(\sigma_{n,a,o}^{i,j}-1\right)\ln V_{n,a,o}^{i,j}+\left(\lambda_{n,a,o}^{i}-1\right)\ln\left(1-V_{n,a,o}^{i,j}\right)
\end{aligned}
\end{equation}
Since only case $\left(z_{n,\tau-1}^{k}=i,z_{n,\tau}^{k}=j\right)$ associated with $\ln V_{n,a,o}^{i,j}$, rearrange terms related to it,
\begin{equation}
\begin{aligned}
&\left[\sum_{k,t}\frac{1}{K}\sum_{\tau=1}^{t}q_{n,t}^k(z_{n,\tau-1}^{k}=i,z_{n,\tau}^{k}=j)\mathbb{I}(a_{n,\tau-1}^{k}=a,o_{n,\tau}^{k}=o)\right]\ln V_{n,a,o}^{i,j} \\
&=\left(\sigma_{n,a,o}^{i,j}-1\right)\ln V_{n,a,o}^{i,j} \\
&\rightarrow \sigma_{n,a,o}^{i,j}=1+\sum_{k,t}\frac{1}{K}\sum_{\tau=1}^{t}q_{n,t}^k(z_{n,\tau-1}^{k}=i,z_{n,\tau}^{k}=j)\mathbb{I}(a_{n,\tau-1}^{k}=a,o_{n,\tau}^{k}=o)
\end{aligned}
\end{equation}
For $\ln\left(1-V_{n,a,o}^{i,j}\right)$, all cases $\left(z_{n,\tau-1}^{k}=i,z_{n,\tau}^{k}=m\right)$ for $m>j$ must be considered, 
\begin{equation}
\medmuskip=0mu
\thinmuskip=0mu
\thickmuskip=0mu
\begin{aligned}
&\left[\sum_{m=j+1}^{|\mathcal{Z}_n|}\frac{1}{K}\sum_{k,t}\sum_{\tau=1}^{t}q_{n,t}^k(z_{n,\tau-1}^{k}=i,z_{n,\tau}^{k}=m)\mathbb{I}(a_{n,\tau-1}^{k}=a,o_{n,\tau}^{k}=o)+\text{E}_{q(\alpha)}\left[\alpha_{n,a,o}^{i}\right]-1\right]\ln\left(1-V_{n,a,o}^{i,j}\right) \\
&=\left(\lambda_{n,a,o}^{i,j}-1\right)\ln\left(1-V_{n,a,o}^{i,j}\right) \\
&\rightarrow \lambda_{n,a,o}^{i,j}=\frac{a_{n,a,o}^{i}}{b_{n,a,o}^{i}}+\sum_{m=j+1}^{|\mathcal{Z}_n|}\frac{1}{K}\sum_{k,t}\sum_{\tau=1}^{t}q_{n,t}^k(z_{n,\tau-1}^{k}=i,z_{n,\tau}^{k}=m)\mathbb{I}(a_{n,\tau-1}^{k}=a,o_{n,\tau}^{k}=o)
\end{aligned}
\end{equation}
Where $\text{E}_{q(\alpha)}\left[\alpha_{n,a,o}^{i}\right]=\frac{a_{n,a,o}^{i}}{b_{n,a,o}^{i}}$.

For the update of each $q(\pi_{n,i})$, rearrange components in \ref{equ:pi_update} in terms of the $q(\pi_{n,i})$ distribution,
\begin{equation}
\begin{aligned}
&\frac{1}{K}\sum_{k,t,z_{n,0:t}^{k}}q(z_{n,0:t}^{k})\sum_{\tau=0}^{t}\ln\pi_{n,z_{\tau}}^{k,a_{\tau}}+\ln p(\pi_{n}) \\
&=\frac{1}{K}\sum_{k,t,z_{n,0:t}^{k}}q(z_{n,0:t}^{k})\sum_{\tau=0}^{t}\ln\pi_{n,z_{\tau}}^{k,a_{\tau}}+\ln \prod_{i=1}^{|\mathcal{Z}_n|}\prod_{a=1}^{|\mathcal{A}_n|}\left(\pi_{n,i}^{a}\right)^{\theta_{n,i}^{a}-1} \\
&=\frac{1}{K}\sum_{k,t,z_{n,0:t}^{k}}q(z_{n,0:t}^{k})\sum_{\tau=0}^{t}\ln\pi_{n,z_{\tau}}^{k,a_{\tau}}+ \sum_{i=1}^{|\mathcal{Z}_n|}\sum_{a=1}^{|\mathcal{A}_n|}\left(\theta_{n,i}^{a}-1\right)\ln\pi_{n,i}^{a} \\
&=\sum_{i=1}^{|\mathcal{Z}_n|}\sum_{a=1}^{|\mathcal{A}_n|}\left[\theta_{n,i}^{a}+\frac{1}{K}\sum_{k,t}\sum_{\tau=0}^{t}q_{n,t}^k(z_{n,\tau}^{k}=i)\mathbb{I}(a_{n,\tau}^{k}=a)-1\right]\ln\pi_{n,i}^{a} \\
&=\sum_{i=1}^{|\mathcal{Z}_n|}\sum_{a=1}^{|\mathcal{A}_n|}\left(\phi_{n,i}^{a}-1\right)\ln\pi_{n,i}^{a} \\
&=\ln\prod_{i=1}^{|\mathcal{Z}_n|}\prod_{a=1}^{|\mathcal{A}_n|}\left(\pi_{n,i}^{a}\right)^{\phi_{n,i}^{a}-1} \\
&=\ln q(\pi_{n}) \\
&\rightarrow \phi_{n,i}^{a}=\theta_{n,i}^{a}+\frac{1}{K}\sum_{k,t}\sum_{\tau=0}^{t}q_{n,t}^k(z_{n,\tau}^{k}=i)\mathbb{I}(a_{n,\tau}^{k}=a)
\end{aligned}
\end{equation}
In the optimal formulas for $q(\Theta_n)$, we need $q_{n,t}^k(z_{n,\tau}^{k})=\Tilde{\nu}_t^k p(z_{n,\tau}^{k}|a_{n,0:t}^{k},o_{n,1:t}^{k},\Tilde{\Theta})$, where $p(z_{n,\tau}^{k}|a_{n,0:t}^{k},o_{n,1:t}^{k},\Tilde{\Theta})$ is the marginal distribution of $p(z_{n,0:t}^{k}|a_{n,0:t}^{k},o_{n,1:t}^{k},\Tilde{\Theta})$. Obtaining it by directly marginalizing the following joint distribution is extremely computationally cumbersome, 
\begin{equation}
\begin{aligned}
    & p(z_{n,\tau}^{k}|a_{n,0:t}^{k},o_{n,1:t}^{k},\Tilde{\Theta}) \\
    & =\sum_{z_{n,\forall t\neq\tau}^{k}=1}^{|\mathcal{Z}_n|}p(z_{n,0:t}^{k}|a_{n,0:t}^{k},o_{n,1:t}^{k},\Tilde{\Theta}) \\
    & =\sum_{z_{n,\forall t\neq\tau}^{k}=1}^{|\mathcal{Z}_n|}\frac{p(a_{n,0:t}^{k},z_{n,0:t}^{k}|o_{n,1:t}^{k},\Tilde{\Theta})}{p(a_{n,0:t}^{k}|o_{n,1:t}^{k},\Tilde{\Theta})} \\
    & p(a_{n,0:t}^{k}|o_{n,1:t}^{k},\Tilde{\Theta})=\sum_{z_{n,0:t}^{k}=1}^{|\mathcal{Z}_n|}p(a_{n,0:t}^{k},z_{n,0:t}^{k}|o_{n,1:t}^{k},\Tilde{\Theta})
\end{aligned}
\end{equation}
Instead, each marginal distribution for $\tau=0,...,t$ can be computed analytically by iterative method. The marginal distribution for each $\tau$ can be factorized into two independent sections according to the d-separation property of Bayes network,
\begin{equation}
\begin{aligned}
    & p(z_{n,\tau}^{k}=i|a_{n,0:t}^{k},o_{n,1:t}^{k},\Tilde{\Theta}) \\
    & \propto p(a_{n,0:t}^{k},z_{n,\tau}^{k}=i|o_{n,1:t}^{k},\Tilde{\Theta}) \\
    & =p(a_{n,0:\tau}^{k},z_{n,\tau}^{k}=i|o_{n,1:\tau}^{k},\Tilde{\Theta})p(a_{n,\tau+1:t}^{k}|z_{n,\tau}^{k}=i,a_{n,\tau:t}^k,o_{n,\tau+1:t}^{k},\Tilde{\Theta}) \\
    & =\alpha_{n,\tau}^k(i)\beta_{n,\tau}^{k,t}(i),
\end{aligned}
\end{equation}
where $\alpha$ and $\beta$ are similar to the forward-backward messages in hidden Markov models. For notational simplicity, we remove $\Tilde{\Theta}$ in the derivation of $\alpha$ and $\beta$. The $\alpha$ and $\beta$ can be computed recursively via dynamic programming,
\begin{equation}
\begin{aligned}
& \alpha_{n,\tau}^k(i) = p(a_{n,0:\tau}^{k},z_{n,\tau}^{k}=i|o_{n,1:\tau}^{k},\Tilde{\Theta}) \\
    & =
    \begin{cases}
     \eta_{n}^i\pi(a_{n,0}^k|z_{n,0}^k=i) & \tau = 0 \\
     \sum_{j=1}^{|\mathcal{z}_n|}\alpha_{n,\tau-1}^k(j)\omega(z_{n,\tau}^k=i|z_{n,\tau-1}^k=j,a_{n,\tau-1}^k,o_{n,\tau}^k)\pi(a_{n,\tau}^k|z_{n,\tau}^k=i) & \tau > 0
    \end{cases}
\end{aligned}
\end{equation}
\begin{equation}
\medmuskip=1mu
\thinmuskip=1mu
\thickmuskip=1mu
\begin{aligned}
& \beta_{n,\tau}^{k,t}(i) = p(a_{n,\tau+1:t}^{k}|z_{n,\tau}^{k}=i,a_{n,\tau}^{k},o_{n,\tau+1:t}^{k},\Tilde{\Theta}) \\
    & =
    \begin{cases}
     1 & \tau = t \\
     \sum_{j=1}^{|\mathcal{z}_n|}\omega(z_{n,\tau+1}^k=j|z_{n,\tau}^k=i,a_{n,\tau}^k,o_{n,\tau+1}^k)\pi(a_{n,\tau+1}^k|z_{n,\tau+1}^k=j)\beta_{n,\tau+1}^{k,t}(j) & \tau < t
    \end{cases}
\end{aligned}
\end{equation}
So the marginal distributions in $q(\Theta)$ update are computed by
\begin{align}
& p(z_{n,\tau}^k=i|a_{n,0:t}^k,o_{n,1:t}^k,\Tilde{\Theta})=\frac{\alpha_{n,\tau}^k(i)\beta_{n,\tau}^{k,t}(i)}{\sum_{i=1}^{|\mathcal{z}_n|}\alpha_{n,\tau}^k(i)\beta_{n,\tau}^{k,t}(i)} \\
&\begin{aligned}
    & p(z_{n,\tau-1}^k=i,z_{n,\tau}^k=j|a_{n,0:t}^k,o_{n,1:t}^k,\Tilde{\Theta}) \\ &=\frac{\alpha_{n,\tau-1}^k(i)\omega(z_{n,\tau}^k=j|z_{n,\tau-1}^k=i,a_{n,\tau-1}^k,o_{n,\tau}^k)\pi(a_{n,\tau}^k|z_{n,\tau}^k=j)\beta_{n,\tau}^{k,t}(j)}{\sum_{i,j=1}^{|\mathcal{z}_n|}\alpha_{n,\tau-1}^k(i)\omega(z_{n,\tau}^k=j|z_{n,\tau-1}^k=i,a_{n,\tau-1}^k,o_{n,\tau}^k)\pi(a_{n,\tau}^k|z_{n,\tau}^k=j)\beta_{n,\tau}^{k,t}(j)}
\end{aligned}
\end{align}

For the update of $q(\rho_n)$, start with the optimal formula and treat all components unrelated to $\rho_n$ as constant,
\begin{equation}
\medmuskip=1mu
\thinmuskip=1mu
\thickmuskip=1mu
\begin{aligned}
&q(\rho_n) \\
&\propto \exp\Biggl\{\text{E}_{q(\Theta,\alpha,z)}\left[\frac{1}{K}\sum_{k,t,z_{n,0:t}^{k}}\ln\Tilde{r}_{t}^{k} p\left(a_{n,0:t}^{k},z_{n,0:t}^{k}\vert o_{n,1:t}^{k},\Tilde{\Theta}\right)\right] \\
&+\text{E}_{q(\Theta,\alpha,z)}\left[\ln p(\Theta\vert\alpha_n, \rho_n)\right]+\text{E}_{q(\Theta,\alpha,z)}\left[\ln p(\rho_n)\right]\Biggr\} \\
&\propto \exp\left\{\text{E}_{q(\Theta,\alpha,z)}\left[\ln p(\Theta\vert\alpha_n, \rho_n)\right]+\text{E}_{q(\Theta,\alpha,z)}\left[\ln p(\rho_n)\right]\right\} \\
&\propto \exp\left\{\text{E}_{q(u)}\left[\ln p(u_n\vert\rho_n)\right]+\ln p(\rho_n)\right\} \\
&=\exp\left\{\text{E}_{q(u)}\left[\ln \prod_{i=1}^{|\mathcal{Z}_n|}p(u_n^i\vert\rho_n)\right]\right\}p(\rho_n) \\
&=\exp\left\{\text{E}_{q(u)}\left[\sum_{i=1}^{|\mathcal{Z}_n|}\ln \frac{\Gamma(1+\rho_n)}{\Gamma(1)\Gamma(\rho_n)}{u_n^i}^{1-1}(1-u_n^i)^{\rho_n-1}\right]\right\}\frac{f^e}{\Gamma(e)}\rho_n^(e-1)\exp\left\{-f\rho_n\right\} \\
&\propto \exp\left\{\sum_{i=1}^{|\mathcal{Z}_n|}\ln\rho_n+(e-1)\ln\rho_n+(\rho_n-1)\sum_{i=1}^{|\mathcal{Z}_n|}\text{E}_{q(u)}\left[\ln(1-u_n^i)\right]-f\rho_n\right\} \\
&=\exp\left\{(e+|\mathcal{Z}_n|-1)\ln\rho_n+(\rho_n-1)\sum_{i=1}^{|\mathcal{Z}_n|}\left[\Psi\left(\mu_n^i\right)-\Psi\left(\delta_n^i+\mu_n^i\right)\right]-f\rho_n\right\} \\
&=\rho_n^{-1}\exp\left\{(e+|\mathcal{Z}_n|)\ln\rho_n-\rho_n\left(f-\sum_{i=1}^{|\mathcal{Z}_n|}\left[\Psi\left(\mu_n^i\right)-\Psi\left(\delta_n^i+\mu_n^i\right)\right]\right)+C_{\rho_n}\right\} \\
&\approx \text{Gamma}(g_n,h_n)
\end{aligned}
\end{equation}
Since $q(\rho_n)$ is assumed to be Gamma distribution, compare the above expression with $\text{Gamma}(g_n,h_n)$, we can obtain
\begin{equation}
\begin{aligned}
& g_n = e+|\mathcal{Z}_n| \\
& h_n = f-\sum_{i=1}^{|\mathcal{Z}_n|}\left[\Psi\left(\mu_n^i\right)-\Psi\left(\delta_n^i+\mu_n^i\right)\right]
\end{aligned}
\end{equation}

For the update of each $q(\alpha_{n,a,o}^{i})$, from optimal formula we have
\begin{equation}
\medmuskip=1mu
\thinmuskip=1mu
\thickmuskip=1mu
\begin{aligned}
&q(\alpha_{n}) \\
&\propto \exp\Biggl\{\text{E}_{q(\Theta,\rho,z)}\left[\frac{1}{K}\sum_{k,t,z_{n,0:t}^{k}}\ln\Tilde{r}_{t}^{k} p(a_{n,0:t}^{k},z_{n,0:t}^{k}\vert o_{n,1:t}^{k},\Theta)\right] \\
&+\text{E}_{q(\Theta,\rho,z)}\left[\ln p(\Theta_n\vert\alpha_n)\right]+\text{E}_{q(\Theta,\rho,z)}\left[\ln p(\alpha_n)\right]\Biggr\} \\
&\propto \exp\left\{\text{E}_{q(\Theta,\rho,z)}\left[\ln p(\Theta_n\vert\alpha_n)\right]+\text{E}_{q(\Theta,\rho,z)}\left[\ln p(\alpha_n)\right]\right\} \\
&\propto\exp\left\{\text{E}_{q(V)}\left[\ln p(V_n\vert\alpha_n)\right]+\ln p(\alpha_n)\right\} \\
&=\exp\left\{\text{E}_{q(V)}\left[\ln p(V_n\vert\alpha_n)\right]\right\}p(\alpha_n) \\
&\text{(for each $\alpha$ with }(n,a,o,i)\text{ indices)} \\
&\rightarrow \exp\left\{\text{E}_{q(V)}\left[\ln\prod_{j=1}^{|\mathcal{Z}_n|}p(V_{j}\vert\alpha)\right]\right\}p(\alpha) \\
&=\exp\left\{\sum_{j=1}^{|\mathcal{Z}_n|}\text{E}_{q(V)}\left[\ln p(V_{j}\vert\alpha)\right]\right\}p(\alpha) \\
&=\exp\left\{\sum_{j=1}^{|\mathcal{Z}_n|}\text{E}_{q(V)}\left[\ln\frac{\Gamma(1+\alpha)}{\Gamma(1)\Gamma(\alpha)}V_j^{1-1}(1-V_j)^{\alpha-1}\right]\right\}\frac{d^{c}}{\Gamma(c)}\alpha^{c-1}\exp\left\{-d\alpha\right\} \\
&\propto \exp\left\{\sum_{j=1}^{|\mathcal{Z}_n|}\ln\alpha+(\alpha-1)\sum_{j=1}^{|\mathcal{Z}_n|}\text{E}_{q(V)}\left[\ln (1-V_j)\right]+(c-1)\ln\alpha-d\alpha\right\} \\
&=\exp\left\{\left(c+|\mathcal{Z}_n|-1\right)\ln\alpha+(\alpha-1)\sum_{j=1}^{|\mathcal{Z}_n|}\left[\Psi(\lambda_j)-\Psi(\sigma_j+\lambda_j)\right]-d\alpha\right\} \\
&=\alpha^{-1}\exp\left\{\left(c+|\mathcal{Z}_n|\right)\ln\alpha+\alpha\left(d-\sum_{j=1}^{|\mathcal{Z}_n|}\left[\Psi(\lambda_j)-\Psi(\sigma_j+\lambda_j)\right]\right)-C_{\alpha}\right\} \\
&\approx \text{Gamma}(a,b)
\end{aligned}
\end{equation}
Similar to $q(\rho_n)$, $q(\alpha)$ is Gamma distribution with parameters $(a,b)$, apply the above optimal formula to all $q(\alpha_{n,a,o}^i)$, the derivation of each $q(\alpha_{n,a,o}^i)$ can be obtained by the following update,
\begin{equation}
\begin{aligned}
& a_{n,a,o}^{i}=c_{n,a,o}+|\mathcal{Z}_n| \\
& b_{n,a,o}^{i}=d_{n,a,o}-\sum_{j=1}^{|\mathcal{Z}_n|}\left[\Psi\left(\lambda_{n,a,o}^{i,j}\right)-\Psi\left(\sigma_{n,a,o}^{i,j}+\lambda_{n,a,o}^{i,j}\right)\right]
\end{aligned}
\end{equation}

\end{document}